\documentclass[letterpaper]{article}
\usepackage[preprint]{aaai2027}
\usepackage{fontawesome5}
\usepackage[hyphens]{url}
\usepackage{graphicx}
\usepackage{natbib}
\usepackage{caption}
\usepackage{booktabs}
\usepackage{multirow}
\usepackage{array}
\usepackage{amsmath}
\usepackage{amssymb}
\usepackage{pifont}
\usepackage{placeins}
\usepackage{algorithm}
\usepackage{algpseudocode}
\usepackage{float}
\newcommand{\cmark}{\ding{51}}
\newcommand{\xmark}{\ding{55}}
\newcommand{\ptmark}{$\triangle$}
\graphicspath{{paper/}{./}}
\makeatletter
\def\input@path{{paper/}{./}}
\makeatother

\title{SciFigAlign: Scoring Scientific Figures by Fine-tuned Alignment of Visuals with Manuscript Evidence}

\newcommand{\projectpagelink}{%
    \leavevmode
    \pdfstartlink
    attr{/Border [0 0 0]}
    user{/Subtype /Link /A << /S /URI /URI (https://github.com/FrankDengAI/SciFigAlign) >>}%
    \faGithub\hspace{0.35em}\textit{Project Page}%
    \pdfendlink
}

\author{
    Chuanzhi Xu$^{1\dagger}$\thanks{Equal contribution.
    \quad $\dagger$ Corresponding authors. \\$\dagger$ Chuanzhi Xu (chuanzhi.xu@sydney.edu.au)\\ $\dagger$ Zihan Deng (zhdeng@hku.hk)},
    Zihan Deng$^{2\dagger}$\footnotemark[1],
    Huiqi Liang$^{1}$,\\
    Chengkun Yue$^{3}$,
    Zhanlin Cui$^{4}$,
    Pengfei Ye$^{5}$,
    Weidong Cai$^{1}$
}

\affiliations{
    $^1$ The University of Sydney
    $^2$ The University of Hong Kong
    $^3$ Indiana University\\
    $^4$ The University of New South Wales
    $^5$ Massachusetts Institute of Technology\\
    \projectpagelink
}

\copyrighttext{}

\begin{document}

\maketitle

\begin{abstract}
Scientific figure assessment in peer review differs fundamentally from general image quality evaluation: a figure must be visually legible, faithfully support the manuscript's claims, and communicate evidence with a clear visual hierarchy. However, if we apply traditional image assessment methods to scientific figure quality assessment, limitations emerge: classic IQA models capture perceptual quality or aesthetics but cannot judge whether a figure serves the paper's scientific argument; CLIP-based methods assess generic image-text correspondence, yet lack understanding of manuscript context; and zero-shot LLM/VLM judges, when repurposed for figure scoring, often yield overly concentrated scores with limited fusion of visual and textual evidence. We introduce an annotated dataset of 3,857 scientific figures from peer-reviewed conference papers, each rated along four peer-review-oriented dimensions: Clarity, Relevance, Informativeness, and Structure. We propose \textbf{SciFigAlign}, a fine-tuned multimodal scorer that grounds figure quality assessment in manuscript evidence. Given a figure crop, caption, citing paragraphs, and light paper context, SciFigAlign fine-tunes CLIP and SciBERT end-to-end with per-modality cross-attention and CubeMLP fusion, jointly optimizing SmoothL1 regression with a within-paper ranking hinge loss. Under paper-level splits, SciFigAlign achieves a macro MAE of \textbf{0.3524} and a within-paper pairwise accuracy of \textbf{81.64\%} on the test set ($n{=}396$), a \textbf{59\% relative error reduction} over the best LLM-as-judge baseline (MAE 0.864). Ablations confirm that manuscript-grounded inputs, citing-context denoising, and ranking supervision are all critical, showing that scientific figure assessment requires learned alignment between visual content and manuscript evidence rather than prompting alone, even with state-of-the-art VLMs.
\end{abstract}

\section{Introduction}

\begin{figure*}[t]
  \centering
  \includegraphics[width=\textwidth]{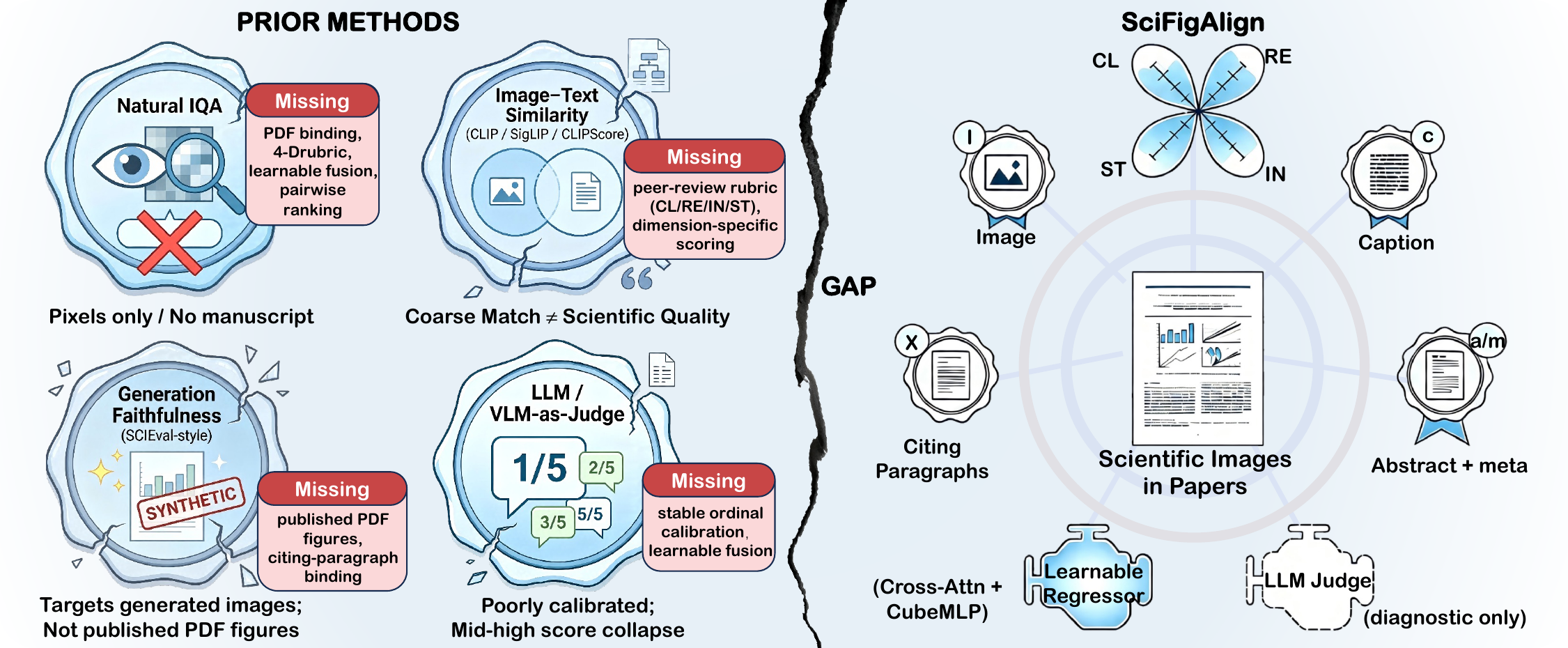}
  \caption{Overview of SciFigAlign versus prior image assessment methods.
    Left: four prior paradigms and what each lacks for scientific figure scoring.
  Right: SciFigAlign binds a published figure to caption, citing paragraphs, and abstract/metadata, scores Clarity / Relevance / Informativeness / Structure (CL/RE/IN/ST) with a learnable CrossAttn+CubeMLP regressor, and keeps an LLM judge as a diagnostic baseline only.}
  \label{fig:overview}
\end{figure*}

\begin{table*}[t]
  \centering
  \small
  \setlength{\tabcolsep}{4pt}
  \begin{tabular}{@{}lcccc@{}}
    \toprule
    \textbf{Comparison dimension} & \textbf{IQA / CLIP} & \textbf{SCIEval-style} & \textbf{LLM-judge} & \textbf{SciFigAlign (Ours)} \\
    \midrule
    Real figures from published CS papers & \ptmark & \xmark & \ptmark & \cmark \\
    Caption + citing-paragraph binding & \xmark & \ptmark & \ptmark & \cmark \\
    Peer-review style multi-dim.\ rubric & \xmark & \cmark & \cmark & \cmark \\
    Learnable multimodal fusion regressor & \ptmark & \cmark & \xmark & \cmark \\
    Within-paper ranking objective & \xmark & \xmark & \ptmark & \cmark \\
    Exportable attention / modality weights & \xmark & \ptmark & \xmark & \cmark \\
    \midrule
    Primary target & pixels / similarity & generated fidelity & zero-shot ordinal & published-figure quality \\
    \bottomrule
  \end{tabular}
  \caption{Capability comparison against related scoring paradigms.
    \cmark=fully supported; \ptmark=partial; \xmark=not targeted.}
  \label{tab:positioning}
\end{table*}

A single well-designed figure can largely impact a paper's empirical narrative, and yet assessing its quality is far more complex than evaluating an everyday photograph. Scientific figures must remain readable at print scale, stay faithful to surrounding claims, and present evidence with a clear visual hierarchy~\cite{borkin2013,wang2004ssim}.
In peer review, a figure is rarely judged on its own. Reviewers always ask whether labels are legible, whether the visual supports the caption and citing text, whether the plot is informative for the claimed contribution, and whether panels admit a coherent reading path.
Automatic scoring methods that disregard these requirements offer limited value for review assistants, authoring tools, or corpus-level analysis of scientific papers.

Natural-image quality is largely perceptual~\cite{brisque2012}: a landscape photograph can often be directly judged from sharpness, color, and aesthetics.
However, scientific figures are different: their utility depends on print-scale legibility, caption fidelity, and honest representation of cited evidence.
If axes are truncated while the text asserts ``consistent gains,'' or if an architecture schematic is decorative rather than informative for the claimed method, pixel-level inspection alone cannot expose the defect.
Assessing scientific paper figures is therefore multimodal: it depends on the crop, the caption, and what the manuscript asserts in citing paragraphs.

Figure~\ref{fig:overview} and Table~\ref{tab:positioning} summarize this shortfall when we attempt to apply general image assessment methods to scientific figure assessment tasks.
Natural Image Quality Assessment (IQA) cannot read the manuscript and cannot detect the visual mismatch of the claim~\cite{brisque2012,bylinskii2017}.
Reference-free similarity (CLIPScore, SigLIP) measures generic image-text matching, but not peer-review Clarity or Informativeness~\cite{clipscore2021,siglip2023}.
Learnable faithfulness models such as SCIEval target generated scientific images~\cite{scieval2026}, but not authentic conference figures already embedded in PDFs with captions and in-text citations.
Zero-shot LLM/VLM judges need no labeled data~\cite{geval2023,prometheus2024,zheng2023mtbench}, yet are hard to calibrate on ordinal scientific rubrics, often concentrate in mid-to-high buckets, and give little control over how visual and textual evidence are fused~\cite{zheng2023mtbench,prabhu2025}.
In short, general image assessment methods should not be directly applied to scientific figure assessment.

Inspired by how peer reviewers compare figures from the same paper side by side, we propose \textbf{SciFigAlign}, a \emph{fine-tuned multimodal scorer} that grounds scientific figure quality assessment aligned with manuscript evidence. We cast scientific figure quality assessment as supervised multimodal regression that targets not merely absolute scoring but relative ordering within each manuscript. Since the process of evaluating images by actual reviewers is essentially about comparing good and bad rather than just giving an independent score, we believe that independent scores have no meaning. Given a published crop with its caption, citing paragraphs, and light paper context, SciFigAlign predicts Clarity, Relevance, Informativeness, and Structure against a 1–5 rubric, jointly optimizing absolute score regression with within-paper ranking.
Our contributions include:
\begin{itemize}
  \item We collect a rubric-annotated corpus dataset of 3,857 figures from peer-reviewed conference papers (1,982 human-rated) with paper-level splits, each bound to its caption and citing text.
  \item We propose SciFigAlign, a multi-stream multimodal fusion model, and a joint SmoothL1 regression $+$ within-paper ranking training objective that couples absolute calibration with relative ordering, together with carefully designed zero-shot LLM baselines for comparison.
  \item With extensive experiments, SciFigAlign achieves 0.3524 MAE and 81.64\% within-paper pairwise accuracy -- a 59\% relative error reduction over the best LLM judge.
\end{itemize}

\section{Related Work}
\label{sec:related}
\noindent \textbf{Traditional Image Quality Assessment.}
Classic IQA methods score blur, noise, and color artifacts on natural photos~\cite{brisque2012,bylinskii2017,rosenholtz2007,wang2004ssim}.
Multimodal judges such as Q-Align can read text printed inside an image~\cite{wu2024qalign}, but they are not trained with external manuscript paragraphs.
If we apply these pipelines to scientific figures, we implicitly assume that visual clarity equals peer-review utility; our rubric rejects this because claim--visual mismatch can only be resolved by reading the paper.
Recent visualization judges~\cite{visjudge2026,prometheus2024} extend rubric-based scoring to charts or generic images, yet still omit index-resolved citing paragraphs and within-paper ranking.
In sum, these lines provide perceptual or template-based baselines rather than manuscript-grounded assessment of whether a published figure supports its claim in context.

\noindent\textbf{Scientific Figure Quality Assessment.}
In 2026, some concurrent research pay closer attention to evaluating scientific images and figures. \textbf{SIQA}~\cite{siqa2026} defines Knowledge (Scientific Validity, Scientific Completeness) and Perception (Cognitive Clarity, Disciplinary Conformity) axes on standalone scientific images.
Its pipeline splits into SIQA-U (multiple-choice understanding) and SIQA-S (alignment with expert quality judgments), but neither stage binds a crop to index-resolved citing paragraphs in a published conference PDF, nor targets within-paper figure ranking under a peer-review rubric.
\textbf{SIU2A}~\cite{siu2a2026} instead builds a forensic loop on \emph{controlled corruptions} (Detail Distortion, Incompleteness, False Content, Entity Confusion), scoring Utility (error detection and repair instructions) and Upgradability (whether an edit restores validity).
That pipeline optimizes diagnosis and repair of edited images, not continuous scoring of authentic figures already embedded in accepted papers.
\textbf{AIBench}~\cite{aibench2026} evaluates \emph{generated} academic illustrations: it derives four-level VQA from method-section logic diagrams to test visual--logical consistency and uses VLMs for aesthetics, rather than regressing multidimensional quality on published PDF crops with full manuscript evidence.
At the system level, no prior work fully covers SciFigAlign's setting -- scoring authentic paper figures with manuscript context, continuous CL/RE/IN/ST scores, and same-paper orderings -- and the closest concurrent lines remain preprint-only and closed-source at the time of this work.
Most nearby work instead targets scientific image generation~\cite{imagereward2023,genfig12026,paperbanana2026,autofigure2026} or text-only manuscript-quality review~\cite{scieval2026,geval2023}, not multidimensional assessment of figures already present in peer-reviewed PDFs.
Appendix~A analyzes these related lines in detail.

\section{Benchmark and Methodology}
\label{sec:method}


\begin{figure*}[t]
  \centering
  \includegraphics[width=\textwidth]{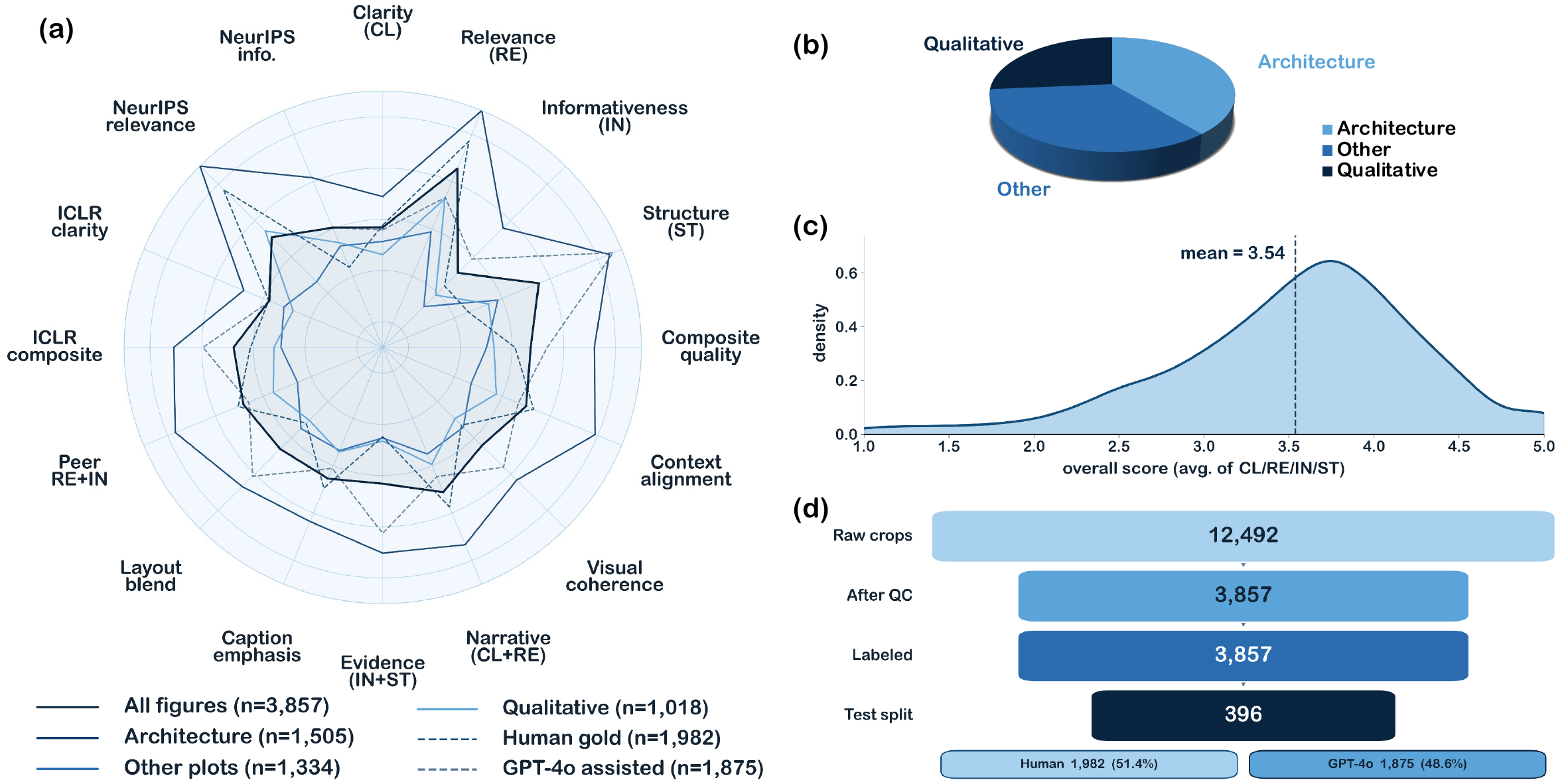}
  \caption{Corpus statistics of dataset in SciFigAlign ($N{=}3{,}857$).
    (a)~Multi-axis rubric profiles by figure type and annotation source.
  (b)~Type mix: architecture 39.0\% / other 34.6\% / qualitative 26.4\%.
  (c)~Overall-score density (mean $3.54$).
  (d)~Construction funnel from 12,492 raw crops to 3,857 labeled figures (test $n{=}396$; human 51.4\% / GPT-4o 48.6\%).}
  \label{fig:data}
\end{figure*}

\begin{table}[t]
  \centering
  \footnotesize
  \setlength{\tabcolsep}{3.5pt}
  \begin{tabular}{@{}cl@{}}
    \toprule
    \textbf{Dim.} & \textbf{1 $\rightarrow$ 3 $\rightarrow$ 5 (condensed)} \\
    \midrule
    CL & Illegible $\rightarrow$ readable w/ issues $\rightarrow$ fully clear \\
    RE & Unrelated $\rightarrow$ partially relevant $\rightarrow$ highly consistent \\
    IN & Negligible content $\rightarrow$ moderate $\rightarrow$ rich \& effective \\
    ST & Chaotic path $\rightarrow$ basically reasonable $\rightarrow$ clear flow \\
    \bottomrule
  \end{tabular}
  \caption{Scoring rubric used by human annotators.}
  \label{tab:rubric}
\end{table}

\subsection{Task Definition and Formulation}
\label{sec:task}

The quality of scientific figures is semantic, not a low-level pixel property~\cite{borkin2013}: a figure must be legible, speak to the surrounding text, have genuine scientific content, and hold together as a unified whole. 

Therefore, we propose the following criteria for assessing scientific figures. \textbf{Clarity ($\mathrm{CL}$)}: whether the figure is clear, readable, and cleanly laid out (fonts, contrast, legends, axis labels). A figure may be scientifically dense yet still score poorly here if clutter or rendering artefacts obstruct reading.
\textbf{Relevance ($\mathrm{RE}$)}: whether the figure matches the caption, citing paragraphs, and the paper's experimental narrative. High-relevance figures do argumentative work; low-relevance ones feel incidental or decorative.
\textbf{Informativeness ($\mathrm{IN}$)}: how much the figure contributes to the paper's argument (comparisons, curves, method details vs.\ removable schematics).
\textbf{Structure ($\mathrm{ST}$)}: subfigure organization, reading path, and visual hierarchy.

We represent each figure instance as a tuple $(I,c,\mathcal{X},a,m)$ (figure crop, caption, citing paragraphs, abstract, metadata), anchored to a source PDF so that $\mathcal{X}$ is index-resolved rather than recovered by layout heuristics alone.
The model predicts continuous scores $\hat{\mathbf{y}}=[\hat{y}_{\mathrm{CL}},\hat{y}_{\mathrm{RE}},\hat{y}_{\mathrm{IN}},\hat{y}_{\mathrm{ST}}]^{\top}$ on the 1--5 rubric scale.

\subsection{Designed Annotation Rubric}
\label{sec:rubric}
We design a human-based annotation rubric for labeling a scientific figure dataset. Annotators score each dimension on a discrete 1--5 scale (Table~\ref{tab:rubric}; full descriptors in Appendix~B).
The rubric follows scientific-communication practice~\cite{borkin2013}, was refined in annotator discussion, and was applied after training with batch-consistency checks.
Separate dimensions make it possible to see \emph{where} a figure fails (legibility, contextual fit, scientific depth, or layout) instead of collapsing everything into one score.

\begin{table}[t]
  \centering
  \small
  \setlength{\tabcolsep}{4pt}
  \begin{tabular}{@{}ll@{}}
    \toprule
    \textbf{Property} & \textbf{SciFigAlign corpus} \\
    \midrule
    Source venues & ICLR / NeurIPS / ICML \\
    Raw extractions & 12,492 figure crops \\
    Final scale & 3,857 figures from 3,126 papers \\
    Annotation sources & 1,982 human + 1,875 GPT-4o-assisted \\
    Test set & 396 human-rated figures \\
    Split protocol & Paper-level 80/10/10 split \\
    Rubric & CL / RE / IN / ST, 1--5 scale \\
    Type mix & Arch. 39.0\%, Other 34.6\%, Qual. 26.4\% \\
    Score distribution & Mean 3.54; concentrated around 3--4 \\
    Context coverage & $>$99\% with caption/citing context \\
    \bottomrule
  \end{tabular}
  \caption{Summary of the SciFigAlign corpus dataset.}
  \label{tab:data}
\end{table}

Overall score is $y=\tfrac{1}{4}\sum_{d\in\mathcal{D}}s(d)$ with $\mathcal{D}=\{\mathrm{CL},\mathrm{RE},\mathrm{IN},\mathrm{ST}\}$; we report macro MAE and Spearman SRCC against $\hat{y}$.
Pairwise accuracy on same-paper pairs with gap at least $\tau$ is defined as:
\begin{equation}
\mathrm{PA}=\frac{1}{|\mathcal{P}|}\sum_{(i,j)\in\mathcal{P}}
\mathbf{1}\big[\mathrm{sign}(\hat{y}_i-\hat{y}_j)=\mathrm{sign}(y_i-y_j)\big],
\label{eq:pa}
\end{equation}
where $y_i$ and $\hat{y}_i$ are overall (mean) scores and $\mathcal{P}=\{(i,j): \mathrm{paper}(i)=\mathrm{paper}(j),\ |y_i-y_j|\ge\tau\}$.

PA complements MAE by evaluating whether a model preserves the relative ordering of figures in the same paper.

\subsection{Dataset Construction and Statistics}
\label{sec:data}

We construct a manuscript-grounded corpus dataset from peer-reviewed machine learning conference papers (Table~\ref{tab:data}; Figure~\ref{fig:data}; complete details in Appendix~B).
Figure crops and captions are extracted from PDFs with PyMuPDF~\cite{pymupdf2024}, while citing paragraphs are recovered by index matching over expressions such as ``Figure~$k$'' and ``Fig.~$k$''.
Mentions of the same figure index are merged within section windows, and overly long windows are truncated for stable batching.
Index-resolved binding is preferred over layout-only heuristics because crops, captions, and in-text mentions are often separated across columns or pages, making nearby OCR text unreliable for recovering $\mathcal{X}$.

Annotations follow the rubric in Table~\ref{tab:rubric}, with human-gold and GPT-4o-assisted labels produced under the same schema after pilot consistency screens.
Paper-level splits prevent leakage across figures from the same manuscript.
The mid--high score concentration summarized in Table~\ref{tab:data} motivates reporting PA alongside macro MAE and SRCC, since MAE alone can reward conservative predictors that fail to recover within-paper orderings.
High context coverage ensures that Relevance and Informativeness are usually evaluable from manuscript evidence.

\subsection{ SciFigAlign Architecture}
\label{sec:arch}

\begin{figure*}[t]
  \centering
  \includegraphics[width=\textwidth]{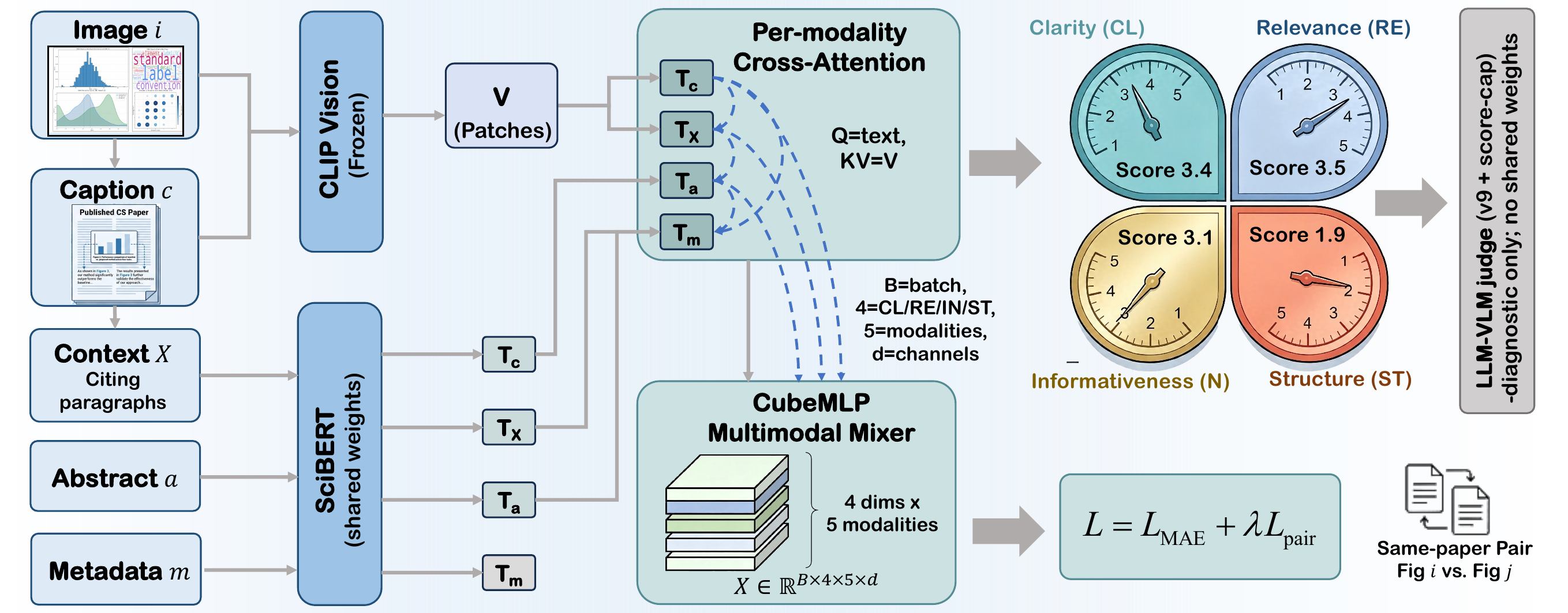}
  \caption{SciFigAlign architecture with a diagnostic LLM/VLM-judge side path.
  Five inputs (image, caption, citing paragraphs, abstract, metadata) are encoded by CLIP Vision and shared SciBERT into patch tokens~$V$ and text tokens~$T_c,T_{\mathcal{X}},T_a,T_m$; per-modality cross-attention uses text as query and~$V$ as key/value; CubeMLP mixes a $B{\times}S{\times}5{\times}d$ tensor (short sequence $S{=}4$ $\times$ five modalities).
  The main path outputs Clarity / Relevance / Informativeness / Structure and is trained with $\mathcal{L}=\mathcal{L}_{\mathrm{reg}}+\lambda\mathcal{L}_{\mathrm{pair}}$ on same-paper pairs; the LLM/VLM judge (v9) is diagnostic only and shares no weights.}
  \label{fig:model}
\end{figure*}

Figure~\ref{fig:model} shows the pipeline.
SciFigAlign encodes five streams (image, caption, context, abstract, metadata), aligns each text stream to visual patches with per-modality cross-attention, fuses with CubeMLP over sequence, modality, and channel axes, and regresses the four quality dimensions~\cite{clip2021,scibert2019,cubemlp2022}.
Unlike frozen embedding ridges, CLIP ViT-B/32 and SciBERT are fine-tuned end-to-end; cross-attention, CubeMLP, and score heads are trained from scratch.
The design goal is explicit control: each text evidence source can attend to different image regions before a shared mixer combines them for the four score heads.

\noindent \textbf{Encoders.}
CLIP ViT maps the image to patch tokens $V\in\mathbb{R}^{N\times d}$ with shared width $d{=}256$.
Each text stream $t\in\{c,\mathcal{X},a,m\}$ is encoded by SciBERT~\cite{scibert2019}, mean-pooled over non-padding tokens, and linearly projected to $T_t\in\mathbb{R}^{d}$.
SciBERT fits scientific terminology better than generic BERT.
The five streams are processed independently at this stage, so later fusion can reweight visual evidence against caption claims, citing context, and paper-level background without forcing an early concatenation of all text fields.

\noindent \textbf{Per-modality Cross-attention.}
Instead of concatenating all text with the image, each text stream queries visual patches separately~\cite{lu2019vilbert}:
\begin{equation}
\tilde{T}_t=\mathrm{CrossAttn}(Q{=}T_t,\ K{=}V,\ V{=}V),
\end{equation}
with $h{=}4$ heads.
This keeps caption wording from being mixed with body citations in a single shared alignment, so each evidence source can form its own correspondence to image regions.
For example, captions can attend to labels and module boxes, citing context to claim-supporting regions, and abstract/metadata to figure type and paper background.
Attention maps are exported for inspection (Appendix~D).
The aligned image stream and four text streams form a length-5 modality sequence for CubeMLP fusion.

\noindent \textbf{CubeMLP Fusion.}
Aligned features form $X\in\mathbb{R}^{B\times S\times 5\times d}$ (batch, short pooled sequence $S{=}4$, five modalities, channels); four score heads are read out after fusion (Appendix~C).

Three stacked CubeMLP blocks apply residual axis-specific MLPs~\cite{cubemlp2022}:
\begin{equation}
X \leftarrow X + \mathrm{MLP}_{\mathrm{axis}}(X),\quad
\mathrm{axis}\in\{\mathrm{seq},\mathrm{mod},\mathrm{chn}\}.
\end{equation}
Sequence/modality mixing models interactions among the five streams; dimension mixing lets the four score heads share related cues (e.g., Clarity and Structure both depend on layout); channel mixing recombines feature dimensions.
Compared with heavy self-attention over the full tensor, this design is lighter and more stable under our labeled-set size.
CubeMLP gates yield global and per-dimension modality-importance weights at inference, which we use as a coarse audit of which streams drive each score.

\noindent \textbf{Score Heads and Context Denoising.}
Four two-layer MLP heads output $\hat{\mathbf{y}}\in[1,5]^4$.
Heads share the fused CubeMLP representation but keep separate parameters, so one dimension can emphasise visual layout while another emphasises claim--text consistency.
In the \emph{denoised} setting, we down-weight weakly related cites (bare ``see Fig.~$k$'') and filter header/footer OCR artefacts in citing windows before encoding.
Without this step, long citing windows often dilute the signal that Relevance and Informativeness need. A complete formalization of the forward computation, including multi-head cross-attention, CubeMLP axis mixing, score-head readout, citing-context denoising, and end-to-end training/inference pseudocode, is provided in Appendix~C.

\subsection{Training Objective}
\label{sec:train}

Figure quality is both an absolute score and a relative judgment within a paper.
Human raters often find pairwise comparison more natural than placing every figure on an absolute scale, especially when most published figures occupy a mid--high quality range.
Accordingly, SciFigAlign combines pointwise regression with a within-paper ranking objective.

\noindent \textbf{Trainable Components.}
Unlike frozen-feature baselines such as SigLIP2-Ridge, or prompt-only LLM judges, SciFigAlign fine-tunes the CLIP ViT-B/32 image encoder and the SciBERT text encoder end-to-end.
The per-modality cross-attention layers, CubeMLP fusion blocks, and four score heads are trained from scratch.
This adapts generic image--text representations to peer-review-oriented dimensions such as print-scale Clarity and claim--visual Relevance.

\noindent \textbf{Pointwise Regression.}
For each figure, the model predicts four rubric scores,
$\hat{\mathbf{y}}\in[1,5]^4$.
Pointwise supervision uses a dimension-averaged SmoothL1 loss:
\begin{equation}
\mathcal{L}_{\mathrm{reg}}
=
\frac{1}{4}
\sum_{d}
\mathrm{SmoothL1}(\hat{y}_d-y_d).
\end{equation}

\noindent \textbf{Within-paper Ranking.}
Because the label distribution is concentrated around mid--high scores, regression alone can encourage conservative predictions that look reasonable under MAE but fail to identify the stronger figure within the same manuscript.
We therefore add a margin-based ranking term over same-paper pairs whose gold-score gap exceeds a threshold:
\begin{equation}
\mathcal{L}_{\mathrm{pair}}
=
\mathbb{E}_{(i,j)}
\Big[
\max
\big(
0,\,
m-\mathrm{sign}(y_i-y_j)(\hat{y}_i-\hat{y}_j)
\big)
\Big].
\end{equation}
The final objective is:
\begin{equation}
\mathcal{L}
=
\mathcal{L}_{\mathrm{reg}}
+
\lambda
\mathcal{L}_{\mathrm{pair}}.
\end{equation}
Full derivations, training/inference pseudocode, hyperparameters, and checkpoint selection are given in Appendix~C.

\section{Experiments and Results}
\label{sec:experiments}

\subsection{Experimental Setup}
\label{sec:setup}

All main experiments are evaluated on a held-out human-rated test subset ($n{=}396$ figures) from the full corpus (Table~\ref{tab:data}), using paper-level 80/10/10 partitioning so that no figures from the same manuscript appear in both training and test.
When supported, methods receive the same manuscript-grounded fields: figure crop, caption, citing paragraphs, abstract, and metadata.
We compare against three baseline families: constant Mean/Median predictors; frozen or non-task-specific similarity models, including CLIPScore~\cite{clipscore2021}, BERTScore~\cite{bertscore2020}, handcrafted visual Ridge, and SigLIP2-Ridge with frozen SigLIP2-SO400M features~\cite{siglip2023}; and prompt-only LLM/VLM judges under the same four-dimensional schema, including GPT-5.4-mini with the full-context v9 prompt and Gemini-2.5-Flash on a 200-figure subset.

We report macro MAE, Spearman SRCC, and within-paper pairwise accuracy (PA; Eq.~\ref{eq:pa}), with 10,000-sample bootstrap confidence intervals for MAE.
SciFigAlign follows the objective in Section~\ref{sec:train}: CLIP ViT-B/32 and SciBERT are fine-tuned end-to-end, while cross-attention, CubeMLP, and four score heads are trained from scratch.
The deployed checkpoint uses denoised inputs and SmoothL1 regression plus the within-paper ranking hinge with $\lambda{=}0.2$, margin $m{=}1.0$, and pair threshold $\tau{=}0.5$.
Full optimization details, hyperparameters, hardware, checkpoint selection, and LLM-judge post-processing are provided in Appendix~C.

\subsection{Main Results}
\noindent \textbf{Overall Performance.} Table~\ref{tab:main} reports overall metrics.
SciFigAlign attains macro MAE \textbf{0.3524}, SRCC 0.3088, and PA \textbf{81.64\%} on 365 valid pairs -- 59\% relative MAE reduction vs.\ LLM-as-judge (0.864) and 60\% vs.\ SigLIP2-Ridge (0.877).
Bootstrap 95\% CI: $[0.331,0.374]$ vs.\ $[0.841,0.887]$ for the judge ($p{<}0.001$).
Constant predictors (MAE $0.897$--$0.937$) cannot rank within a paper (PA $0\%$).
Among non-learned methods, LLM-as-judge is strongest on absolute MAE (0.864) but only reaches PA $63.2\%$ and macro SRCC $0.198$; SigLIP2-Ridge ranks slightly better (PA $65.8\%$, SRCC $0.266$) despite worse MAE.
Similarity baselines (CLIPScore / BERTScore) stay near chance on pairwise order (${\approx}53$--$55\%$ PA).
SciFigAlign improves all three metrics jointly, with the largest relative gain on PA ($81.64\%$).

\begin{table}[t]
  \centering
  \small
  \setlength{\tabcolsep}{3.5pt}
  \begin{tabular}{@{}lccc@{}}
    \toprule
    Method & MAE$\downarrow$ & SRCC$\uparrow$ & PA$\uparrow$ \\
    \midrule
    Mean predictor & 0.937 & -- & 0.0\% \\
    Median predictor & 0.897 & -- & 0.0\% \\
    CLIPScore$^{\dagger}$ & 0.891 & 0.154 & 55.3\% \\
    BERTScore$^{\dagger}$ & 0.903 & 0.079 & 52.6\% \\
    Visual Ridge$^{\dagger}$ & 0.884 & 0.191 & 58.1\% \\
    SigLIP2-Ridge & 0.877 & 0.266 & 65.8\% \\
    LLM-as-judge (v9) & 0.864 & 0.198 & 63.2\% \\
    \underline{\textbf{SciFigAlign}} & \underline{\textbf{0.3524}} & \underline{\textbf{0.3088}} & \underline{\textbf{81.64\%}} \\
    \bottomrule
  \end{tabular}\\[-2pt]
  \caption{Overall test metrics.
    MAE: mean of four per-dimension MAEs.
    Baseline SRCC: macro-average of per-dimension Spearman; SciFigAlign SRCC is on overall $\hat{y}$.
    PA: within-paper pairs with gap ${\ge}0.5$ (Eq.~\ref{eq:pa}).
    $^{\dagger}$Smaller human-rated subset ($n{=}185$); all other rows use the main test subset ($n{=}396$). Constant predictors: SRCC undefined; PA $0\%$ under Eq.~\ref{eq:pa}.}
  \label{tab:main}
\end{table}


\begin{figure}[t]
  \centering
  \includegraphics[width=\linewidth]{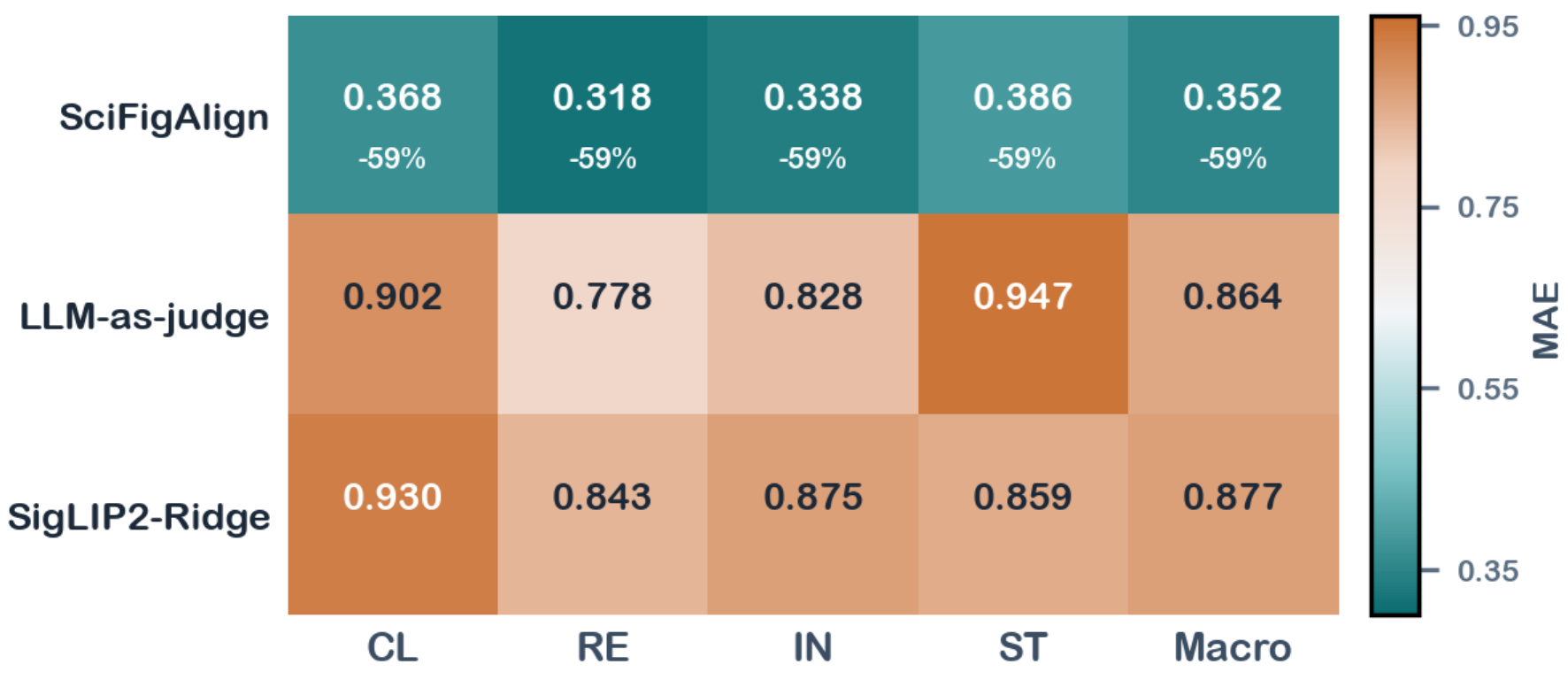}
  \caption{Per-dimension test MAE heatmap.}
  \label{fig:main-results}
\end{figure}

\noindent \textbf{Per-dimension and Category Breakdown.} Figure~\ref{fig:main-results} breaks errors by dimension. SciFigAlign is best on every dimension, with fairly even relative gains.
Relevance is the easiest head for SciFigAlign (MAE $0.318$), consistent with caption--cite binding providing a strong supervisory signal; Structure remains hardest ($0.386$), reflecting irregular multi-panel layouts that are also difficult for humans.
The LLM judge is stronger on Informativeness (MAE 0.828) than Structure (0.947) or Clarity (0.902), suggesting fluent prose cues help more on content density than on layout or print-scale legibility -- exactly where a fine-tuned visual stream should help.
SigLIP2-Ridge has its best SRCC on Structure ($0.391$), consistent with a strong visual backbone, yet MAE stays near $0.86$--$0.93$ across dimensions: similarity features correlate with layout somewhat but do not calibrate ordinal scores.
Architecture diagrams are easiest for the judge; qualitative grids and heterogeneous ``other'' plots are hardest (structure MAE up to ${\approx}1.05$ in exploratory type breakdowns).
Residual SciFigAlign errors concentrate on OCR-dense charts and Informativeness cases where the visual is sparse, but the caption claims a strong result.

\noindent \textbf{Ranking vs.\ Absolute Error.}
PA $81.64\%$ on 365 valid same-paper pairs shows that ranking-aware training recovers relative order far better than a mid-score constant would.
SRCC stays moderate ($0.3088$): ordinal correlation on compressed labels is harder than gap-filtered pairwise accuracy, so we report both.
For triage (e.g., flagging the weaker figure in a paper), PA is the more actionable metric; MAE remains needed for absolute score reporting.
The gap between PA and SRCC is therefore not a contradiction: many within-paper pairs have clear winners under $\tau{=}0.5$, while global rank correlation is diluted by near-ties around the mid--high mode.

\subsection{Ablations and Analysis}
\begin{figure}[t]
  \centering
  \includegraphics[width=\linewidth]{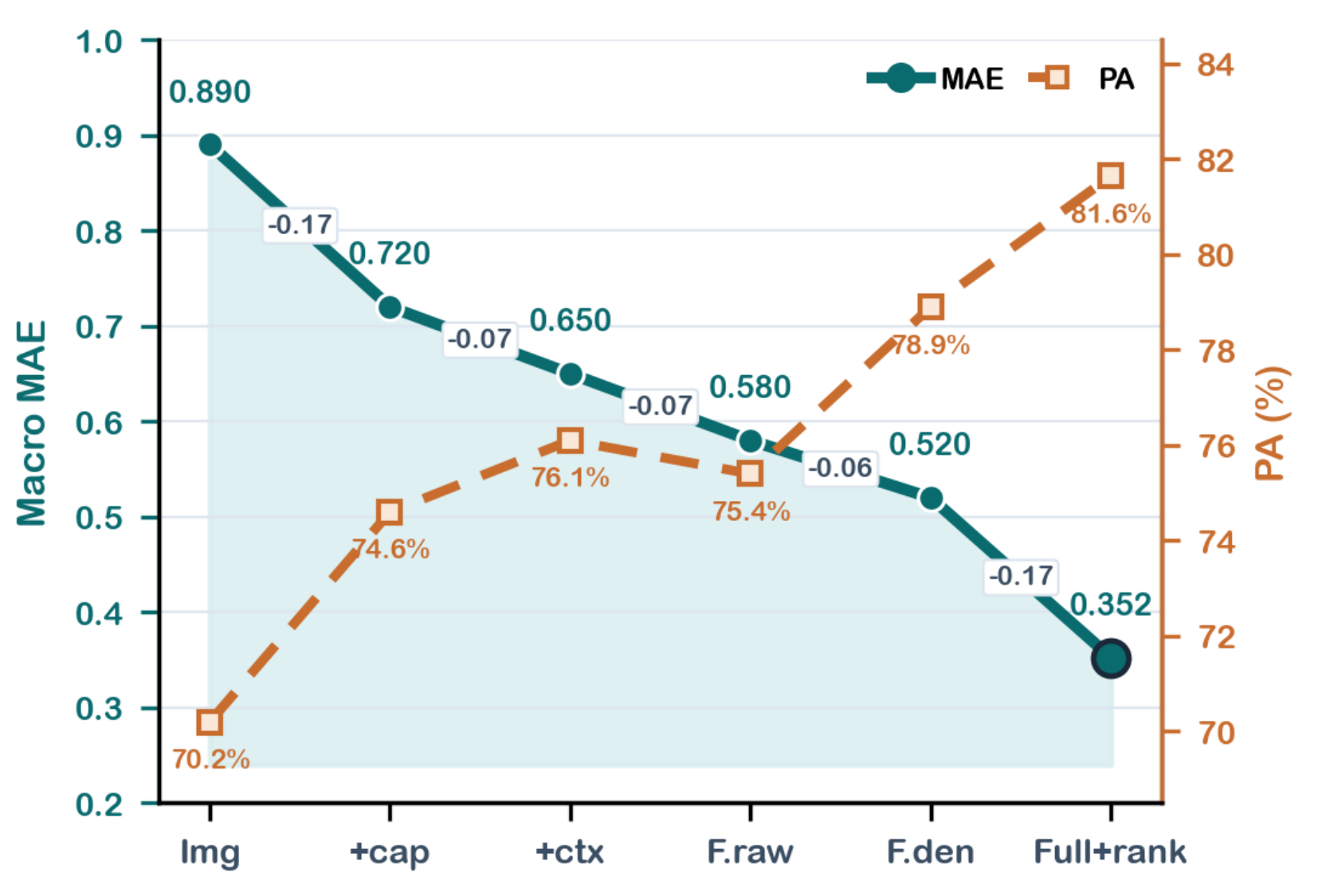}
  \caption{Test ablation progression ($n{=}396$).}
  \label{fig:ablation}
\end{figure}

Table~\ref{tab:ablation} and Figure~\ref{fig:ablation} vary inputs and ranking under identical paper-level splits.
On the test ladder, caption and context cut MAE ($0.890\to0.720\to0.650$); denoising beats raw full text ($0.520$ vs.\ $0.580$); Full+rank reaches \textbf{0.3524} MAE and \textbf{81.64\%} PA.
The largest steps are caption addition and ranking (${\approx}{-}0.17$ MAE each); PA rises from $70.2\%$ to $81.6\%$.
A matched validation protocol finds Image+Caption strongest on absolute MAE ($0.2733$) while Full Denoised$+\lambda{=}0.10$ peaks Spearman ($0.5011$). Absolute calibration and ranking prefer slightly different $\lambda$; we report both and deploy the checkpoint chosen by validation loss ($0.0999$).

\begin{table}[t]
  \centering
  \small
  \setlength{\tabcolsep}{3.5pt}
  \begin{tabular}{@{}lcccc@{}}
    \toprule
    Configuration & MAE$\downarrow$ & SRCC$\uparrow$ & PA$\uparrow$ & Split \\
    \midrule
    Image only & 0.890 & 0.439 & 70.2\% & test \\
    + caption & 0.720 & 0.488 & 74.6\% & test \\
    + context & 0.650 & 0.498 & 76.1\% & test \\
    Full raw & 0.580 & 0.472 & 75.4\% & test \\
    Full denoised & 0.520 & 0.497 & 78.9\% & test \\
    Full+rank (val) & 0.2733 & 0.5011 & 80.1\% & val \\
    \underline{\textbf{Full+rank (test)}} & \underline{\textbf{0.3524}} & \underline{\textbf{0.3088}} & \underline{\textbf{81.64\%}} & \underline{\textbf{test}} \\
    \bottomrule
  \end{tabular}
  \caption{Ablations (test ladder + val Full+rank).
    Val row is not equated to test.}
  \label{tab:ablation}
\end{table}

\noindent \textbf{Multimodal Fine-tuning and Ranking.}
The ladder isolates three failure modes of prior pipelines: visual-only scoring, noisy full-text concatenation, and regression-only training on compressed mid--high labels.
Fine-tuning CLIP/SciBERT with task heads adapts generic encoders to peer-review dimensions; the hinge recovers within-paper order beyond what MAE alone suggests.
Caption addition alone removes nearly as much test error as the final ranking term, underscoring that published-figure quality is not a unimodal visual property.
Denoising beats raw full text by $0.06$ MAE: boilerplate figure pointers dilute the citing stream unless down-weighted, so more context is not automatically better.
Matched splits keep these effects attributable to input and objective choices rather than encoder swaps.
The val/test MAE gap ($0.2733$ vs.\ $0.3524$) is expected under different selection criteria and should not be read as a single-number claim of $0.27$ on the canonical test set.

\noindent \textbf{Baselines and Judge Ceilings.}
Similarity ridges and constant predictors remain near or above 0.85 MAE, and constants yield $0\%$ PA.
The LLM judge is the strongest non-learned baseline on absolute error, but uneven across dimensions and only mid-60s on within-paper PA ($63.2\%$): it benefits from fluent caption language on Informativeness while remaining weak on Structure, Clarity, and relative ordering under mid--high score concentration.
SigLIP2-Ridge is slightly stronger on ranking (PA $65.8\%$) than the LLM judge, consistent with its higher macro SRCC, yet still far below SciFigAlign.
SciFigAlign improves uniformly (Figure~\ref{fig:main-results}): shared visual tokens queried by each text stream appear to help all four heads rather than trading one dimension for another.
On a 200-subset, GPT-5.4-mini (0.778) beats Gemini-2.5-Flash (0.874) but stays near an always-4 constant (0.800), illustrating MAE ceiling effects on 3--4-dominated labels.
Text-only and image-only judges underperform full context, yet even full-context prompting remains far from SciFigAlign on the held-out test set (Appendix~F).
Prompt engineering and isotonic post-processing therefore raise the judge floor, but do not replace supervised multimodal calibration.

\noindent \textbf{Attention and Modality Weights.}
Attention maps typically highlight axis labels and method boxes rather than whitespace; modality weights put context and abstract near $0.21$ and image near $0.18$ (Appendix~D).
Citing context and abstract-level narrative contribute beyond the image stream, while the image remains necessary for Clarity and Structure.
Exportable weights and maps make the model more auditable than a black-box LLM verdict, without replacing human review.
Qualitatively, captions often attend to legend and title regions, while citing context focuses on claim-supporting panels -- behavior consistent with the per-modality CrossAttn design.

\noindent \textbf{Error Modes.}
High-error cases often involve crowded multi-panel qualitative grids, OCR-dense tables rendered as figures, and Informativeness disagreements when the visual is sparse but the caption claims a strong result.
Structure errors often coincide with irregular panel order; Clarity errors with tiny axis labels that humans penalize at print scale but patch pooling may under-resolve.
Stronger OCR/layout features for the visual stream are a natural next step, while keeping the ranking objective for within-paper triage.

\noindent \textbf{Discussion.}
Image-only MAE $0.890$ is more than double the full multimodal result, so scoring crops without captions or cites is systematically weak.
Even a carefully engineered LLM judge (MAE $0.864$) stays far above SciFigAlign; adding pixels or text to the prompt does not close the gap.
Under mid--high label concentration, the ranking term is among the two largest MAE steps on the test ladder and lifts PA to $81.64\%$.
Published-figure quality assessment therefore benefits from three ingredients that prior pipelines rarely combine: (i)~manuscript-bound multimodal inputs, (ii)~learned fusion with dimension-specific heads, and (iii)~an explicit within-paper ranking objective.
For practical triage -- choosing which figure to revise in a submission -- relative order matters as much as absolute MAE, which is why we treat PA as a first-class metric alongside regression error.

\section{Conclusion}

We introduce \textbf{SciFigAlign}, a unified framework for scientific figure quality assessment through manuscript-grounded multimodal fusion and within-paper ranking, accompanied by a rubric-annotated benchmark of 3,857 figures from peer-reviewed CS conference PDFs for systematic evaluation.
Our experiments reveal that state-of-the-art LLM/VLM judges struggle to calibrate continuous Clarity/Relevance/Informativeness/Structure scores on authentic PDF crops, and that within-paper ordering gains are fundamentally constrained by citing-context binding and denoising rather than prompt engineering alone.
Through SciFigAlign, we expose critical bottlenecks in existing pipelines, highlighting the gap between isolated-image perceptual scoring and manuscript-grounded peer-review utility.
This framework provides a foundation for ranking and triaging figures within a submission, and is extendable to OCR-dense plots, enhanced Informativeness grounding, and broader venues requiring auditable visual–text alignment.

\bibliography{scifigalign}

\clearpage
\appendix
\noindent{\centering\Large\bfseries Supplementary Appendix\par}
\vspace{0.8em}
\setcounter{table}{0}
\setcounter{figure}{0}
\setcounter{algorithm}{0}
\renewcommand{\thetable}{\arabic{table}}
\renewcommand{\thefigure}{\arabic{figure}}
\renewcommand{\thealgorithm}{\arabic{algorithm}}

\noindent{\textbf{Appendix Guideline: }}This supplement expands protocol material summarized in the main paper:
related-method justification (Appendix~A), full scoring rubric and corpus construction (B),
architecture formalization / pseudocode / training hyperparameters and losses (C),
explainability and dashboards (D; Figures~1--6), training dynamics / corpus / paradigm (E),
extended validation ablations, judge modes, and failures (F), and limitations (G).
The main text refers to these sections by name (Appendix~A--G). \textbf{Overview.}
\textbf{A}~justification vs.\ related scientific-figure assessment methods;
\textbf{B}~full rubric, annotation, and corpus construction;
\textbf{C}~architecture equations and algorithms, hyperparameters, SmoothL1$+$ranking, denoising, LLM-judge;
\textbf{D}~modality/attention signals, dashboards, and scoring cases (Figures~1--6);
\textbf{E}~training dynamics, modality contribution, corpus, paradigm;
\textbf{F}~validation suite, test ladder, judge modes, failures, reproducibility;
\textbf{G}~limitations and scope of claims.

\section{Why SciFigAlign Differs from Related Scientific Figure Assessment Methods}
\label{app:related-justify}

\subsection{Purpose}

To our knowledge, only three recent lines explicitly target \emph{scientific figure/image quality assessment} in a sense close to SciFigAlign: SIQA~\cite{siqa2026}, SIU2A~\cite{siu2a2026}, and AIBench~\cite{aibench2026}.
A reader may therefore ask whether SciFigAlign is simply another variant of these systems.
\textbf{It is not:} they optimize different objects, inputs, pipelines, and success metrics.
This appendix states the boundary paper by paper and documents why we do not reproduce or report head-to-head numbers on their benchmarks.

\subsection{What SciFigAlign Evaluates}

SciFigAlign scores \emph{real figures already published} in peer-reviewed CS conference PDFs (ICLR, NeurIPS, ICML).
For each crop it reads five linked fields---image $I$, caption $c$, all index-resolved citing paragraphs $\mathcal{X}$, abstract $a$, and metadata $m$---and predicts four continuous peer-review scores on a 1--5 rubric: Clarity, Relevance, Informativeness, and Structure.
The model is trained to support a practical workflow: \emph{rank or triage figures within the same paper} when preparing or conducting review.

\subsection{Three Closest Lines and Task Boundaries}
\label{app:three-closest}

Below, each entry follows the same format: \emph{pipeline}, \emph{metric}, \emph{gap vs.\ SciFigAlign}, and \emph{reproducibility status}.

\paragraph{SIQA~\cite{siqa2026}.}
\emph{Pipeline:} SIQA models scientific image quality along Knowledge (Scientific Validity, Scientific Completeness) and Perception (Cognitive Clarity, Disciplinary Conformity).
SIQA-U tests semantic comprehension through multiple-choice tasks; SIQA-S aligns model outputs with expert quality judgments on the SIQA Challenge benchmark of standalone scientific images (e.g., molecular structures, reaction schematics, geometric diagrams).
\emph{Metric:} MCQ accuracy (SIQA-U) and expert-score alignment (SIQA-S).
\emph{Gap:} figures are evaluated as detached scientific images, not as in situ conference-PDF crops with index-bound citing paragraphs; the protocol does not implement manuscript-grounded CL/RE/IN/ST regression or within-paper ranking.
\emph{Reproducibility:} at the time of writing, SIQA does \emph{not} publicly release its benchmark assets or training/evaluation code, so we cannot replicate SIQA-U/S on our corpus.

\paragraph{SIU2A~\cite{siu2a2026}.}
\emph{Pipeline:} SIU2A (Scientific Image Utility and Upgradability Assessment) builds controlled corruptions under four types---Detail Distortion, Incompleteness, False Content, and Entity Confusion---and runs a two-stage protocol: Utility (error detection, localization, repair-instruction generation) and Upgradability (whether an edit restores scientific validity).
\emph{Metric:} structured error understanding and correction quality on manipulated images across biology, chemistry, and physics.
\emph{Gap:} a forensic repair loop on edited images, not rubric-based scoring of authentic figures already embedded in peer-reviewed CS PDFs.
\emph{Reproducibility:} SIU2A appears only as an arXiv preprint (arXiv:2606.03401) without formal venue acceptance at submission time; we therefore treat it as a related but non-replicable benchmark for our setting.

\paragraph{AIBench~\cite{aibench2026}.}
\emph{Pipeline:} AIBench evaluates \emph{generated} academic illustrations.
It constructs logic graphs from method-section text, derives four hierarchical VQA levels (component existence, local topology, phase architecture, global semantics), and combines VQA-based logical checks with VLM-based aesthetic assessment.
\emph{Metric:} VQA accuracy on ${\sim}5{,}704$ QA pairs over 300 top-conference papers, plus aesthetic scores from model judges.
\emph{Gap:} the object is whether a \emph{synthesized} framework figure matches paper logic and looks acceptable---not whether an \emph{existing} published figure is clear, on-claim, informative, and well structured relative to citing text.
\emph{Reproducibility:} AIBench likewise remains an arXiv preprint (arXiv:2603.28068) without formal publication at submission time, and its pipeline targets generation evaluation rather than our published-figure regression task.

\subsection{Other Adjacent Work (Not Published-Figure Quality Assessment)}

Many nearby papers use ``scientific figure,'' ``scientific image,'' or ``academic illustration'' in their titles, but belong to \emph{different task families}.
We group them explicitly so they are not confused with SciFigAlign's published-figure quality regression.

\noindent\textbf{(i) Text-primary or text--image faithfulness evaluation.}
These pipelines center on \emph{language}---prompts, captions, or manuscript excerpts---rather than scoring existing PDF figure crops under a peer-review rubric.
\textbf{SCIEval}~\cite{scieval2026} trains CLIP and LMM modules to judge whether generated scientific images or captioning outputs faithfully match their corresponding textual descriptions, evaluated across three dimensions (Relevance, Accuracy, and Explainability) on the proposed SCIEval-Bench benchmark. It serves as a unified faithfulness evaluation metric for scientific vision-language generation tasks (covering both text-to-image synthesis and image captioning). Critically, it is not designed for continuous assessment of figures already embedded in peer-reviewed PDFs (i.e., it does not rely on citing-paragraph evidence from surrounding text); instead, it strictly focuses on evaluating model-generated content against given prompts or reference pairs.
\textbf{ImageReward}~\cite{imagereward2023} and related preference models likewise score text-to-image outputs for human alignment, not published conference figures inside PDFs.
More broadly, text-only scientific writing or review evaluation (e.g., LLM-as-judge on prose with NLG-style metrics~\cite{geval2023}) never assigns multidimensional figure-quality scores to published crops and therefore lies outside our task.

\noindent\textbf{(ii) Scientific illustration / image \emph{generation} evaluation.}
These systems judge whether models can \emph{produce} acceptable framework figures or experimental visuals, not whether authors' existing conference figures are review-ready.
Besides AIBench (above), recent lines include \textbf{GENFIG1}~\cite{genfig12026} (Figure-1 generation plus six-dimensional VLM judging of \emph{generated} summaries), \textbf{PaperBanana}~\cite{paperbanana2026} (illustration agents with faithfulness/readability VLM judges), and \textbf{AutoFigure}~\cite{autofigure2026} / FigureBench (publication-ready illustration generation benchmarks).
\textbf{SciFig}-style pipelines automate figure \emph{synthesis} and may attach rubric checks to outputs, but they still evaluate generative products rather than regressing quality on figures already accepted into PDFs.
Their metrics (VQA pass rates, VLM judge scores, prompt alignment) measure generation success, not manuscript-grounded CL/RE/IN/ST regression on accepted PDF figures.

\noindent\textbf{(iii) Figure understanding and chart QA.}
A large literature tests whether models can \emph{read} figures.
\textbf{SciFIBench}~\cite{scifibench2024} (${\sim}2000$ MCQ items from arXiv figure--caption pairs), \textbf{FigureQA}~\cite{kahou2018figureqa}, \textbf{ChartQA}~\cite{chartqa2022}, \textbf{PlotQA}~\cite{plotqa2020}, \textbf{CharXiv}~\cite{wang2024charxiv}, and \textbf{SciGraphQA}~\cite{li2023scigraphqa} report comprehension or reasoning accuracy on plots, charts, or scientific graphs.
\textbf{SPUR}~\cite{spur2026} extends this pattern to wet-lab experimental images (${\sim}4264$ QA pairs over ${\sim}1084$ microscopy/gel-style panels), but remains reasoning QA rather than peer-review rubric regression on CS conference PDF figures.
These benchmarks answer ``what is in the figure?'' or ``which caption matches?'' rather than ``is this figure clear, relevant, informative, and well structured for peer review?''

\noindent\textbf{(iv) Visualization-only or generic judge protocols.}
\textbf{VisJudge-Bench}~\cite{visjudge2026} evaluates MLLMs as judges on standalone \emph{data-visualization} images (Fidelity / Expressiveness / Aesthetics over ${\sim}3090$ chart samples), not mixed architecture/qualitative/plot figures inside ML papers with citing context.
\textbf{Prometheus-Vision}~\cite{prometheus2024} and general LMM IQA judges such as \textbf{Q-Align}~\cite{wu2024qalign} provide rubric-based or level-based scoring templates, but they target natural photos or generic fine-grained judging rather than manuscript-indexed conference-figure regression with within-paper ranking.

\noindent\textbf{(v) Caption generation, extraction, and figure--text alignment.}
\textbf{SciCap}~\cite{scicap2021} and \textbf{SciCap+}~\cite{scicapplus2024} generate or augment scientific figure captions; \textbf{S1-MMAlign}~\cite{wang2026s1mmalign} scales scientific figure--text pairing for understanding pretraining.
FigEx-style systems focus on panel detection and caption alignment extraction.
These pipelines improve description, retrieval, or pairing---they do not output continuous multidimensional peer-review scores on published PDF crops tied to citing paragraphs.

\noindent\textbf{Placement relative to SciFigAlign.}
Only the three lines in \S\ref{app:three-closest} are closest in \emph{naming} to scientific figure/image quality assessment; the works above are adjacent but task-mismatched---text faithfulness, generation judging, figure QA, visualization judging, caption/alignment tooling, or generic VLM judges rather than regression on $(I,c,\mathcal{X},a,m)$ with within-paper ranking.

\subsection{Why We Do Not Reproduce These Benchmarks}

None of the three systems above releases a turnkey pipeline that accepts our instance format $(I,c,\mathcal{X},a,m)$ and outputs continuous CL/RE/IN/ST scores with within-paper ranking.
SIQA further lacks public artifacts; SIU2A and AIBench are preprint-only and task-mismatched.
Our main-paper experiments therefore compare only baselines that share the same fields on the held-out human-rated test subset, rather than claiming superiority on SIQA-U, SIU2A-Utility, or AIBench VQA accuracy.

\subsection{Summary}

\begin{table*}[ht]
  \centering
  \footnotesize
  \setlength{\tabcolsep}{3pt}
  \renewcommand{\arraystretch}{1.05}
  \caption{Task-level contrast among the three closest scientific-figure assessment lines and SciFigAlign.}
  \label{tab:related-contrast}
  \resizebox{\textwidth}{!}{%
  \begin{tabular}{@{}lcccc@{}}
    \toprule
    Criterion & SIQA & SIU2A & AIBench & \textbf{SciFigAlign} \\
    \midrule
    Object: published PDF figure & $\times$ & $\times$ & $\times$ & $\checkmark$ \\
    Index-bound citing paragraphs & $\times$ & $\times$ & $\times$ & $\checkmark$ \\
    Continuous peer-review rubric (CL/RE/IN/ST) & $\triangle$ & $\times$ & $\times$ & $\checkmark$ \\
    Primary paradigm & MCQ / judge & corruption repair & gen.\ VQA + aesthetics & regression + PA \\
    Within-paper ranking metric & $\times$ & $\times$ & $\times$ & $\checkmark$ \\
    Public benchmark/code for replication & $\times$ & $\triangle$ & $\triangle$ & \checkmark \\
    \bottomrule
  \end{tabular}}
\end{table*}

\noindent\textbf{Takeaway.}
SIQA, SIU2A, and AIBench are the nearest neighbors in terminology, yet each defines a different problem.
SciFigAlign targets manuscript-grounded, continuous, within-paper assessment of figures already present in peer-reviewed CS PDFs---a workflow none of the three implements end-to-end under publicly reproducible conditions at submission time.

\section{Dataset Construction and Annotation Details}
\label{app:rubric}

\subsection{Rubric Motivation and Design}

The main paper prints a condensed 1--5 rubric (CL/RE/IN/ST); Table~\ref{tab:full-rubric} gives the full annotator-facing descriptors.
Scientific figure quality is semantic rather than a low-level pixel property~\cite{borkin2013}: a figure must be legible, speak to the surrounding text, carry genuine scientific content, and hold together as an organised whole.
These demands motivate Clarity, Relevance, Informativeness, and Structure.
The rubric follows scientific-communication practice~\cite{borkin2013}, was refined in annotator discussion, and was applied after training with batch-consistency checks.
Separate dimensions make failures easier to locate (legibility, contextual fit, scientific depth, or layout).

\begin{table*}[ht]
  \centering
  \small
  \setlength{\tabcolsep}{3pt}
  \renewcommand{\arraystretch}{1.08}
  \caption{Full 1--5 scoring rubric used by human annotators (main paper reports a condensed version).}
  \label{tab:full-rubric}
  \begin{tabular}{@{}c>{\raggedright\arraybackslash}p{0.21\textwidth}>{\raggedright\arraybackslash}p{0.21\textwidth}>{\raggedright\arraybackslash}p{0.22\textwidth}>{\raggedright\arraybackslash}p{0.22\textwidth}@{}}
    \toprule
    Score & Clarity & Relevance & Informativeness & Structure \\
    \midrule
    1 & Most content is illegible & Clearly unrelated to paper & Negligible information content & Chaotic layout, no clear reading path \\
    2 & Partially readable, with difficulty & Weakly related & Limited useful content & Poorly organised \\
    3 & Readable but with noticeable issues & Partially relevant & Moderate information & Basically reasonable structure \\
    4 & Mostly clear, minor defects & Well aligned with text & Informative & Logical hierarchy \\
    5 & Fully clear and easy to read & Highly consistent with paper & Rich and effective scientific content & Clear layout, logical flow \\
    \bottomrule
  \end{tabular}
\end{table*}

\subsection{Annotation Sources and Quality Control}

Human gold labels ($n{=}1{,}982$) store evidence snippets with each score; spans without evidence are revised in QC.
GPT-4o-assisted labels ($n{=}1{,}875$) use the same JSON schema after pilot screens (Krippendorff's $\alpha>0.6$).
Paper-level splits (80/10/10, seed~42) prevent leakage; the held-out test subset uses $n{=}396$ human-rated figures only.

The two annotation sources follow the same four-dimensional rubric and the same score schema.
Human labels are treated as the gold source and are used for the canonical test set, while GPT-4o-assisted labels expand the training corpus under the same annotation format.
The assisted labels are not used as a separate teacher model and are not distilled into SciFigAlign; they are included only as labeled instances that pass the same schema and pilot-consistency requirements.

During annotation, each figure is judged with its available manuscript evidence rather than as an isolated image.
Annotators consider the figure crop, caption, and citing paragraphs when assigning Relevance and Informativeness scores.
This is necessary because a visually polished figure can still fail to support the paper's claim, while a visually simple figure may be highly informative if it directly supports an experimental argument.

\subsection{Corpus Collection and PDF Processing}

SciFigAlign is constructed from peer-reviewed machine learning conference papers from ICLR, NeurIPS, and ICML.
These venues were selected because they provide a large number of publicly available PDFs with diverse scientific figure types, including model architectures, quantitative plots, ablation summaries, and qualitative visualizations.
The goal is to evaluate figures in the context in which they are used by authors, rather than as standalone images.

For each paper, we parse the PDF and extract candidate figure crops and captions using PyMuPDF~\cite{pymupdf2024}.
The initial extraction stage yields 12,492 raw figure crops.
Raw crops include duplicate regions, partial panels, low-quality renderings, and non-figure visual elements.
After deduplication and crop quality control, the final corpus contains 3,857 labeled figures from 3,126 papers.

Each retained figure instance is represented as:
\begin{equation}
(I,c,\mathcal{X},a,m),
\end{equation}
where $I$ is the figure crop, $c$ is the caption, $\mathcal{X}$ is the set of citing paragraphs, $a$ is the abstract, and $m$ denotes lightweight metadata such as venue, year, and figure index when available.

\subsection{Figure--Text Binding}

A central design choice in SciFigAlign is to bind each figure to manuscript evidence through figure-index resolution rather than nearby layout text alone.
For each extracted figure, we recover citing paragraphs by matching explicit references such as ``Figure~$k$'', ``Fig.~$k$'', and common variants.
When the same figure index is mentioned multiple times in the manuscript, we merge the corresponding hosting paragraphs within section-level windows.
Overly long windows are truncated to ensure stable batching during model training.

Index-resolved binding is preferred over layout-only heuristics for two reasons.
First, conference PDFs often place figure crops, captions, and citing paragraphs in different columns or even on different pages.
Second, the text visually surrounding a figure crop may include unrelated paragraphs, headers, footers, or neighboring figure captions.
As a result, relying only on OCR or spatial proximity can assign the wrong evidence to a figure.
By contrast, index matching anchors the context field $\mathcal{X}$ to explicit manuscript references.

The recovered context is particularly important for the Relevance and Informativeness dimensions.
A figure may be visually clear but irrelevant to the claim made in the surrounding text, or it may contain little scientific evidence despite having a polished layout.
Such cases cannot be judged reliably from the crop alone.

\subsection{Deduplication and Crop Quality Control}

The raw extraction stage includes several sources of noise:
\begin{itemize}
    \item duplicate crops caused by repeated PDF objects or overlapping extraction boxes;
    \item partial crops that contain only one panel or a truncated figure region;
    \item non-figure elements such as icons, logos, tables, or decorative graphics;
    \item low-resolution or rendering-corrupted crops;
    \item crops whose figure index cannot be reliably linked to a caption or citing context.
\end{itemize}

We remove exact and near-duplicate crops, discard severely truncated or unreadable regions, and require a usable caption or manuscript anchor.
This filtering reduces the pool from 12,492 raw crops to 3,857 labeled figures.
The resulting corpus retains a broad range of scientific figure types while excluding extraction artifacts that would make annotation unreliable.

\subsection{Score Aggregation and Evaluation Motivation}

For each figure, the overall score is computed as the mean of the four rubric dimensions:
\begin{equation}
y=\frac{1}{4}\sum_{d\in\mathcal{D}}s(d),
\end{equation}
where
\begin{equation}
\mathcal{D}=\{\mathrm{CL},\mathrm{RE},\mathrm{IN},\mathrm{ST}\}.
\end{equation}

The overall score distribution is concentrated around the 3--4 range, with mean 3.54.
This concentration is expected for peer-reviewed conference papers: most published figures are at least usable, while extremely poor or exceptionally polished figures are less frequent.
However, the same concentration creates an evaluation challenge.
A conservative predictor that always outputs a mid-to-high score can obtain a deceptively reasonable MAE while failing to identify which figure in a paper is better or worse.

For this reason, we report pairwise accuracy in addition to macro MAE and Spearman SRCC.
For same-paper pairs with a score gap of at least $\tau$, pairwise accuracy is defined as:
\begin{equation}
\mathrm{PA}=
\frac{1}{|\mathcal{P}|}
\sum_{(i,j)\in\mathcal{P}}
\mathbf{1}
\left[
\mathrm{sign}(\hat{y}_i-\hat{y}_j)
=
\mathrm{sign}(y_i-y_j)
\right],
\end{equation}
where
\begin{equation}
\mathcal{P}=
\{(i,j): \mathrm{paper}(i)=\mathrm{paper}(j), |y_i-y_j|\ge\tau\}.
\end{equation}
This metric evaluates whether a model preserves the relative ordering of figures within the same manuscript, which better matches how reviewers often compare figures during paper assessment.
Here $y_i$ and $\hat{y}_i$ denote the \emph{overall} gold and predicted scores (means over the four dimensions), not the four-dimensional vectors.

\subsection{Splits, Figure Types, and Context Coverage}

We use paper-level splits rather than figure-level random splits.
This is important because multiple figures from the same paper often share visual style, caption conventions, topic vocabulary, and repeated manuscript context.
If figures from the same paper appeared in both training and test sets, models could exploit paper-specific cues rather than learning general figure-quality criteria.

The corpus is split into train, validation, and test partitions with an 80/10/10 ratio using seed~42.
The held-out test subset contains 396 human-rated figures.
Using a human-rated test set ensures that the primary evaluation is grounded in human judgments rather than assisted labels.

Architecture diagrams form the largest group, accounting for 39.0\% of the corpus.
Other diagrams and plots account for 34.6\%, while qualitative figures account for 26.4\%.
This mixture reflects the visual diversity of machine learning papers, where figures may present model architectures, quantitative curves, ablation plots, qualitative examples, or heterogeneous multi-panel summaries.

Context-field coverage exceeds 99\% after filtering.
In practice, this means that almost every retained figure has usable caption and citing-context evidence.
This high coverage is important because Relevance and Informativeness are manuscript-dependent dimensions.
Without context, a model can only infer whether a figure looks plausible; it cannot reliably judge whether the figure supports the paper's actual claim.

The remaining cases without complete context are mostly due to unusual PDF formatting, missing explicit figure references, or extraction artifacts.
We keep only cases with sufficient evidence for reliable scoring.
This design ensures that SciFigAlign evaluates scientific figures as manuscript-grounded objects rather than as standalone images.

\subsection{Corpus Summary}

The final SciFigAlign corpus contains 3,857 labeled figures from 3,126 papers, distilled from 12,492 raw crops.
It includes 1,982 human labels and 1,875 GPT-4o-assisted labels under the same four-dimensional rubric.
The held-out human-rated test subset contains 396 figures.
The corpus uses paper-level 80/10/10 splits, covers architecture, other, and qualitative figure types, and provides over 99\% context coverage with caption and citing-paragraph evidence.
Together, these design choices allow SciFigAlign to evaluate not only visual legibility, but also whether a figure is relevant, informative, and structurally effective in the context of the manuscript.

\section{Architecture Formalization, Training Details, and Judge Protocol}
\label{app:train}
\label{app:arch}

This appendix expands the compact architecture description in the main paper (Section~3.4) into a step-by-step formalization.
We first define inputs and outputs, then describe encoders, cross-attention, CubeMLP fusion, score heads, and citing-context denoising.
We next state the joint SmoothL1$+$within-paper ranking objective and give training/inference pseudocode.
Finally, we list hyperparameters, the checkpoint-selection rule, the compute environment, and the diagnostic LLM-judge protocol.

\subsection{Notation and Problem Instantiation}
\label{app:arch-notation}

Each figure instance is a manuscript-bound tuple
\begin{equation}
z=(I,c,\mathcal{X},a,m),\qquad
\hat{\mathbf{y}}=f_\theta(z)\in\mathbb{R}^{4},
\end{equation}
where $I$ is the figure crop, $c$ the caption, $\mathcal{X}$ the index-resolved citing paragraphs, $a$ the abstract, and $m$ light metadata (venue, year, figure type when available).
The four outputs are Clarity, Relevance, Informativeness, and Structure:
\begin{equation}
\hat{\mathbf{y}}=[\hat{y}_{\mathrm{CL}},\hat{y}_{\mathrm{RE}},\hat{y}_{\mathrm{IN}},\hat{y}_{\mathrm{ST}}]^{\top}.
\end{equation}
Gold labels use the discrete 1--5 rubric; model outputs are continuous scores on the same numeric scale (see Score Heads below).
Let $\mathcal{T}=\{c,\mathcal{X},a,m\}$ be the four text streams and $\mathcal{M}=\{I\}\cup\mathcal{T}$ the five modalities.
In the deployed configuration, the shared projection width is $d{=}256$, CubeMLP uses $L_{\mathrm{cube}}{=}3$ blocks, and cross-attention uses $h{=}4$ heads with $L_{\mathrm{ca}}{=}1$ layer.

\paragraph{Reading guide.}
The pipeline has four stages: (1)~encode image and text independently; (2)~let each text stream attend to image patches; (3)~fuse the five modality streams with CubeMLP; (4)~predict four scores with separate heads.
Algorithms~\ref{alg:forward}--\ref{alg:infer} package the same steps as pseudocode.

\subsection{Encoders}
\label{app:arch-enc}

\paragraph{Vision encoder.}
CLIP ViT-B/32 maps the crop $I$ to patch tokens (the CLS token is discarded):
\begin{align}
V^{(0)} &= \mathrm{CLIP}(I)\in\mathbb{R}^{N\times d_{\mathrm{img}}}, \\
V &= \mathrm{Linear}_{\mathrm{img}}(V^{(0)})\in\mathbb{R}^{N\times d}.
\end{align}
Here $N$ is the number of image patches and $d_{\mathrm{img}}$ is CLIP's native width.
The linear layer projects patches into the shared width $d$.

\paragraph{Text encoder.}
Each text field $t\in\mathcal{T}$ is truncated to at most 256 tokens and encoded by a shared SciBERT:
\begin{equation}
H_t=\mathrm{SciBERT}(t)\in\mathbb{R}^{L_t\times d_{\mathrm{txt}}}.
\end{equation}
We mean-pool over non-padding tokens and apply a modality-specific linear map:
\begin{equation}
u_t=\frac{\sum_{\ell=1}^{L_t}\alpha_{t,\ell}\,H_{t,\ell}}{\sum_{\ell=1}^{L_t}\alpha_{t,\ell}},\qquad
T_t=\mathrm{Linear}_t(u_t)\in\mathbb{R}^{d},
\end{equation}
where $\alpha_{t,\ell}\in\{0,1\}$ is the attention mask.
This matches the main-paper exposition: one $d$-dimensional vector per text stream.
In code, we equivalently keep a short length-$S$ token sequence after adaptive pooling ($S{=}4$) so that cross-attention and CubeMLP operate on a compact sequence; the mathematics below is unchanged if $T_t$ is treated as a length-$1$ sequence.

\subsection{Per-modality Cross-attention}
\label{app:arch-ca}

We do \emph{not} concatenate all text into one query.
Instead, each text stream attends to the same visual patches on its own, so caption wording is not mixed with body citations before alignment.

For text query $Q_t\in\mathbb{R}^{S\times d}$ (or $1\times d$ after mean pooling) and visual keys/values $V\in\mathbb{R}^{N\times d}$, multi-head attention with $h$ heads is
\begin{align}
\mathrm{head}_i
&=
\mathrm{softmax}\!\left(
\frac{(Q_t W_i^{Q})(V W_i^{K})^{\top}}{\sqrt{d_h}}
\right)
(V W_i^{V}),\\
\mathrm{MHA}(Q_t,V)
&=
\mathrm{Concat}(\mathrm{head}_1,\ldots,\mathrm{head}_h)\,W^{O},
\end{align}
with $d_h=d/h$.
A residual feed-forward block then produces the aligned text stream:
\begin{align}
\tilde{Q}_t
&=
\mathrm{LN}\!\big(Q_t+\mathrm{Dropout}(\mathrm{MHA}(Q_t,V))\big),\\
\tilde{T}_t
&=
\mathrm{LN}\!\big(\tilde{Q}_t+\mathrm{Dropout}(\mathrm{FFN}(\tilde{Q}_t))\big).
\end{align}
We export head-averaged attention maps $A_t\in\mathbb{R}^{S\times N}$ for inspection (Appendix~D).
Intuitively: captions often attend to titles and legends; citing paragraphs to claim-supporting panels; abstract/metadata to coarse figure type and paper background.

\subsection{CubeMLP Fusion}
\label{app:arch-cube}

After cross-attention we have five modality streams: a pooled visual stream $V_S=\mathrm{Pool}_{N\to S}(V)$ and four aligned text streams $\tilde{T}_c,\tilde{T}_{\mathcal{X}},\tilde{T}_a,\tilde{T}_m$.
We stack them into a 4D tensor and add a learned stream-type embedding $E\in\mathbb{R}^{5\times d}$:
\begin{equation}
X^{(0)}\in\mathbb{R}^{B\times S\times 5\times d},\qquad
X^{(0)}_{b,s,j,:}=\mathrm{stream}_{b,s,j}+E_{j}.
\end{equation}
Here $B$ is batch size, $S$ is the short sequence length after pooling ($S{=}4$ in our implementation), the third axis indexes the five modalities, and $d$ is the channel width.

\paragraph{Relation to the main-paper tensor notation.}
The main paper writes $X\in\mathbb{R}^{B\times 4\times 5\times d}$ and labels the second axis as ``score dimensions.''
That is a pedagogical shorthand for the same fusion block: with $S{=}4$, CubeMLP mixes a length-$4$ sequence against five modalities, and four dimension-specific queries later read out CL/RE/IN/ST features from the fused tensor.
In other words, the implementation mixes over a short sequence axis of size $4$; the four score heads are applied \emph{after} fusion, not as four independent CubeMLP copies.

Each CubeMLP block applies three residual MLPs, one per axis:
\begin{align}
X &\leftarrow X + \mathrm{MLP}_{\mathrm{seq}}\!\big(\mathrm{LN}(X)\big)
\quad\text{(mix along the sequence axis)},\\
X &\leftarrow X + \mathrm{MLP}_{\mathrm{mod}}\!\big(\mathrm{LN}(X)\big)
\quad\text{(mix along modalities)},\\
X &\leftarrow X + \mathrm{MLP}_{\mathrm{chn}}\!\big(\mathrm{LN}(X)\big)
\quad\text{(mix along channels)}.
\end{align}
Stacking $L_{\mathrm{cube}}$ blocks and a final LayerNorm yields $X^{(L)}$.
Let $\bar{X}\in\mathbb{R}^{B\times 5\times d}$ be the mean of $X^{(L)}$ over the sequence axis.
A small gate network $\mathrm{Gate}$ produces exportable global modality weights
\begin{equation}
\pi=\mathrm{softmax}\big(\mathrm{Gate}(\bar{X})\big)\in\mathbb{R}^{5},\qquad \sum_{j=1}^{5}\pi_j=1.
\end{equation}
For each score dimension $d$, a learned query $q_d\in\mathbb{R}^{d}$ produces dimension-specific modality weights and a pooled feature:
\begin{equation}
\pi_{d,j}=\mathrm{softmax}_j\!\big(\langle\bar{X}_{:,j,:},\,q_d\rangle\big),\qquad
\phi_d=\sum_{j=1}^{5}\pi_{d,j}\,\bar{X}_{:,j,:}.
\end{equation}

\subsection{Score Heads}
\label{app:arch-heads}

Let $g=\mathrm{mean}(X^{(L)})$ be the globally pooled fused vector (mean over sequence and modality axes).
For each dimension $d\in\{\mathrm{CL},\mathrm{RE},\mathrm{IN},\mathrm{ST}\}$,
\begin{equation}
\hat{y}_d
=
5\cdot\sigma\!\big(
\mathrm{MLP}_d([g;\phi_d])
\big),
\end{equation}
where $\sigma$ is the sigmoid and $[g;\phi_d]$ is concatenation.
Thus each head outputs a continuous value in $(0,5)$, which we treat as a soft score on the 1--5 rubric used by annotators.
Separate heads let Clarity/Structure emphasise layout cues while Relevance/Informativeness emphasise claim--text consistency.

\subsection{Citing-context Denoising}
\label{app:arch-denoise}

Long citing windows often contain boilerplate (``see Fig.~$k$'') and OCR header/footer noise.
Before encoding $\mathcal{X}$, we apply a lightweight cleaner $\mathcal{D}$:
\begin{enumerate}
  \item drop or down-weight bare figure pointers that carry no claim-bearing language;
  \item strip repeated header/footer OCR artefacts;
  \item truncate the cleaned window to the 256-token encoder budget.
\end{enumerate}
The deployed checkpoint uses $\mathcal{X}^{\prime}=\mathcal{D}(\mathcal{X})$ (``full denoised'').
Main-paper ablations show that feeding raw long citing text is worse than this filtered stream.

\subsection{Joint Training Objective}
\label{app:arch-loss}

Training has two goals: (i)~predict absolute scores close to the human rubric; (ii)~preserve which figure is better within the same paper.

\paragraph{Pointwise regression.}
With gold labels $\mathbf{y}\in\{1,\ldots,5\}^4$ and residual $r=\hat{y}_d-y_d$,
\begin{equation}
\mathrm{SmoothL1}(r)=
\begin{cases}
\tfrac{1}{2}r^{2}/\beta & |r|<\beta,\\
|r|-\tfrac{1}{2}\beta & \text{otherwise,}
\end{cases}
\qquad(\beta{=}1),
\end{equation}
\begin{equation}
\mathcal{L}_{\mathrm{reg}}
=
\tfrac{1}{4}\sum_{d}\mathrm{SmoothL1}(\hat{y}_d-y_d).
\end{equation}

\paragraph{Within-paper ranking.}
Let the overall gold/prediction be the means
$y=\tfrac{1}{4}\sum_d y_d$ and $\hat{y}=\tfrac{1}{4}\sum_d\hat{y}_d$.
Over same-paper pairs with a clear gap $|y_i-y_j|\ge\tau$, we use a margin hinge:
\begin{equation}
\mathcal{L}_{\mathrm{pair}}
=
\mathbb{E}_{(i,j)}
\Big[
\max\!\big(
0,\,
m-\mathrm{sign}(y_i-y_j)(\hat{y}_i-\hat{y}_j)
\big)
\Big].
\end{equation}
Near-ties ($|y_i-y_j|<\tau$) are dropped because mid--high labels are concentrated and tiny gaps are unstable.

\paragraph{Joint objective.}
\begin{equation}
\mathcal{L}=\mathcal{L}_{\mathrm{reg}}+\lambda\mathcal{L}_{\mathrm{pair}}.
\end{equation}
Deployed settings: $m{=}1.0$, $\tau{=}0.5$, $\lambda{=}0.2$, after sweeping $\lambda\in\{0.05,0.10,0.20\}$ on validation.

\subsection{Pseudocode}
\label{app:arch-algo}

\begin{algorithm}[H]
\caption{SciFigAlign forward pass}
\label{alg:forward}
\begin{algorithmic}[1]
\Require figure tuple $z=(I,c,\mathcal{X},a,m)$; parameters $\theta$
\Ensure scores $\hat{\mathbf{y}}$; optional attention / modality weights
\State $\mathcal{X}' \gets \mathcal{D}(\mathcal{X})$ \Comment{citing-context denoising}
\State $V \gets \mathrm{Linear}_{\mathrm{img}}(\mathrm{CLIP}(I))$ \Comment{image patch tokens}
\For{each text field $t\in\{c,\mathcal{X}',a,m\}$}
  \State $T_t \gets \mathrm{Linear}_t(\mathrm{Pool}(\mathrm{SciBERT}(t)))$
  \State $\tilde{T}_t \gets \mathrm{CrossAttnBlock}(Q{=}T_t,\ K{=}V,\ V{=}V)$
\EndFor
\State Build $X^{(0)}$ from $\{V_S,\tilde{T}_c,\tilde{T}_{\mathcal{X}},\tilde{T}_a,\tilde{T}_m\}$ plus stream embeddings
\For{$\ell=1$ to $L_{\mathrm{cube}}$}
  \State $X^{(\ell)} \gets \mathrm{CubeMLPBlock}(X^{(\ell-1)})$ \Comment{seq / mod / chn MLPs}
\EndFor
\State $X \gets \mathrm{LN}(X^{(L_{\mathrm{cube}})})$;\; $g \gets \mathrm{mean}(X)$
\State Compute modality weights $\pi$ and per-dimension features $\phi_d$
\For{each dimension $d\in\{\mathrm{CL},\mathrm{RE},\mathrm{IN},\mathrm{ST}\}$}
  \State $\hat{y}_d \gets 5\cdot\sigma(\mathrm{MLP}_d([g;\phi_d]))$
\EndFor
\State \Return $\hat{\mathbf{y}}$, exported $\{A_t\}$, $\{\pi$, $\{\pi_d\}$
\end{algorithmic}
\end{algorithm}

\begin{algorithm}[H]
\caption{Same-paper pair mining for ranking}
\label{alg:pairs}
\begin{algorithmic}[1]
\Require minibatch $\mathcal{B}=\{(z_i,\mathbf{y}_i,\mathrm{paper}_i)\}$; threshold $\tau$
\Ensure pair set $\mathcal{P}_{\mathcal{B}}$
\State $\mathcal{P}_{\mathcal{B}} \gets \emptyset$
\For{all unordered pairs $(i,j)$ in $\mathcal{B}$}
  \If{$\mathrm{paper}_i=\mathrm{paper}_j$ and $|y_i-y_j|\ge\tau$}
    \State $\mathcal{P}_{\mathcal{B}} \gets \mathcal{P}_{\mathcal{B}}\cup\{(i,j)\}$
  \EndIf
\EndFor
\State \Return $\mathcal{P}_{\mathcal{B}}$ \Comment{near-ties excluded}
\end{algorithmic}
\end{algorithm}

\begin{algorithm}[H]
\caption{End-to-end SciFigAlign training}
\label{alg:train}
\begin{algorithmic}[1]
\Require train/val splits; hyperparameters $(\lambda,m,\tau,\eta,\ldots)$; seed $s$
\Ensure best checkpoint $\theta^\star$
\State SetSeed($s$); initialize CLIP, SciBERT, CrossAttn, CubeMLP, score heads
\For{epoch $=1$ to $E_{\max}$}
  \For{each minibatch $\mathcal{B}$}
    \State $\{\hat{\mathbf{y}}_i\}_{i\in\mathcal{B}} \gets$ Forward($\{z_i\}$) \Comment{Alg.~\ref{alg:forward}}
    \State $\mathcal{L}_{\mathrm{reg}} \gets \tfrac{1}{4|\mathcal{B}|}\sum_{i,d}\mathrm{SmoothL1}(\hat{y}_{i,d}-y_{i,d})$
    \State $\mathcal{P}_{\mathcal{B}} \gets$ MinePairs($\mathcal{B},\tau$) \Comment{Alg.~\ref{alg:pairs}}
    \If{$\mathcal{P}_{\mathcal{B}}=\emptyset$}
      \State $\mathcal{L}_{\mathrm{pair}} \gets 0$
    \Else
      \State $\mathcal{L}_{\mathrm{pair}} \gets \mathrm{mean}_{(i,j)\in\mathcal{P}_{\mathcal{B}}}\max\!\big(0,\,m-\mathrm{sign}(y_i-y_j)(\hat{y}_i-\hat{y}_j)\big)$
    \EndIf
    \State $\mathcal{L} \gets \mathcal{L}_{\mathrm{reg}}+\lambda\mathcal{L}_{\mathrm{pair}}$
    \State Update $\theta$ with AdamW on $\nabla_\theta\mathcal{L}$ (FP16)
  \EndFor
  \State Evaluate validation loss / MAE / SRCC; early-stop if no improvement
\EndFor
\State \Return $\theta^\star$ with lowest validation loss
\end{algorithmic}
\end{algorithm}

\begin{algorithm}[H]
\caption{Inference and explainability export}
\label{alg:infer}
\begin{algorithmic}[1]
\Require checkpoint $\theta^\star$; figure tuple $z$
\Ensure $\hat{\mathbf{y}}$; modality / attention diagnostics
\State Run Alg.~\ref{alg:forward} with denoising $\mathcal{D}$ enabled
\State Optionally export modality weights and per-stream attention maps
\State \Return scores (e.g., for ranking figures within the same paper)
\end{algorithmic}
\end{algorithm}

\subsection{Training Configuration and Hyperparameters}

\begin{table*}[t]
  \centering
  \small
  \caption{Deployed SciFigAlign training hyperparameters.}
  \label{tab:hparams}
  \setlength{\tabcolsep}{4pt}
  \begin{tabular}{@{}p{3.2cm}p{4.2cm}p{3.2cm}p{4.2cm}@{}}
    \toprule
    Setting & Value & Setting & Value \\
    \midrule
    Image / text encoders & CLIP ViT-B/32 / SciBERT & Freeze encoders & both false (fine-tuned) \\
    Projection width $d$ & 256 & CubeMLP depth / exp. & 3 / 2.0 \\
    Cross-attn heads / layers & 4 / 1 (text$\to$vision) & Dropout / max text & 0.1 / 256 tok. \\
    Batch / epochs & 4 / 5--10 (early stop) & Optimizer / lr / wd & AdamW / $2{\times}10^{-5}$ / $10^{-4}$ \\
    Precision & FP16 & Regression loss & SmoothL1 ($\beta{=}1$) \\
    Ranking $\lambda$ / $m$ / $\tau$ & 0.2 / 1.0 / 0.5 & $\lambda$ sweep & $\{0.05,0.10,0.20\}$ \\
    Split & paper 80/10/10, seed 42 & Best val.\ loss & 0.0999 \\
    Hardware & single GPU, 8GB & Runtime & $\approx$45 min / 5 epochs \\
    Software & PyTorch $\ge$2.0 (FP16) & Checkpoint input & full denoised context \\
    \bottomrule
  \end{tabular}
\end{table*}

Table~\ref{tab:hparams} lists the deployed run that produces the main-paper test numbers.
CLIP ViT-B/32 and SciBERT are fine-tuned end-to-end; cross-attention, CubeMLP, and the four score heads are trained from scratch.
Training uses paper-level 80/10/10 splits with seed~42, batch size~4, FP16, and early stopping within 5--10 epochs.
All text fields are truncated to 256 tokens.
The deployed input setting is full denoised citing context with ranking weight $\lambda{=}0.2$.

\subsection{Checkpoint Selection and $\lambda$ Trade-off}

The loss equations are given in \S\ref{app:arch-loss}; here we only record how the final checkpoint was chosen.
We sweep $\lambda\in\{0.05,0.10,0.20\}$ under matched validation splits.
Absolute calibration and ranking prefer slightly different $\lambda$:
$\lambda{=}0.10$ yields the strongest validation Spearman ($0.5011$), while $\lambda{=}0.2$ with denoised inputs yields the lowest validation loss ($0.0999$) and is the deployed test checkpoint.
For absolute-error ablations, Image+Caption is strongest on validation MAE ($0.2733$), but the final model uses full denoised manuscript evidence plus ranking because that setting best matches the intended within-paper triage use case.

\subsection{Implementation Environment}

Experiments run on a single GPU with 8\,GB memory using PyTorch~$\ge$2.0 and FP16 mixed precision.
A typical 5-epoch pass over the labeled corpus takes about 45 minutes.
Inference can export CubeMLP modality weights and cross-attention maps for diagnostic inspection (Appendix~D).

\subsection{Diagnostic LLM Judge Protocol}

The LLM/VLM judge is a zero-shot diagnostic baseline only.
It shares no weights with SciFigAlign, and its scores are never distilled into the model.

The v9 prompt scores the figure crop, caption, and citing context under the same four-dimensional schema, returning JSON scores and rationales for Clarity, Relevance, Informativeness, and Structure.
It includes figure-type caps and anti-positivity / anti-compensation rules to reduce mid--high score inflation.
Unless noted otherwise, reported judge numbers use uncapped isotonic post-processing on the full-context v9 prompt.
We also evaluate text-only, image-only, image+caption, and full-context variants on a 200-figure subset, including a GPT-5.4-mini vs.\ Gemini-2.5-Flash comparison.
These variants are baselines only and are not used for SciFigAlign training.

\begin{figure*}[t]
  \centering
  \includegraphics[width=\textwidth]{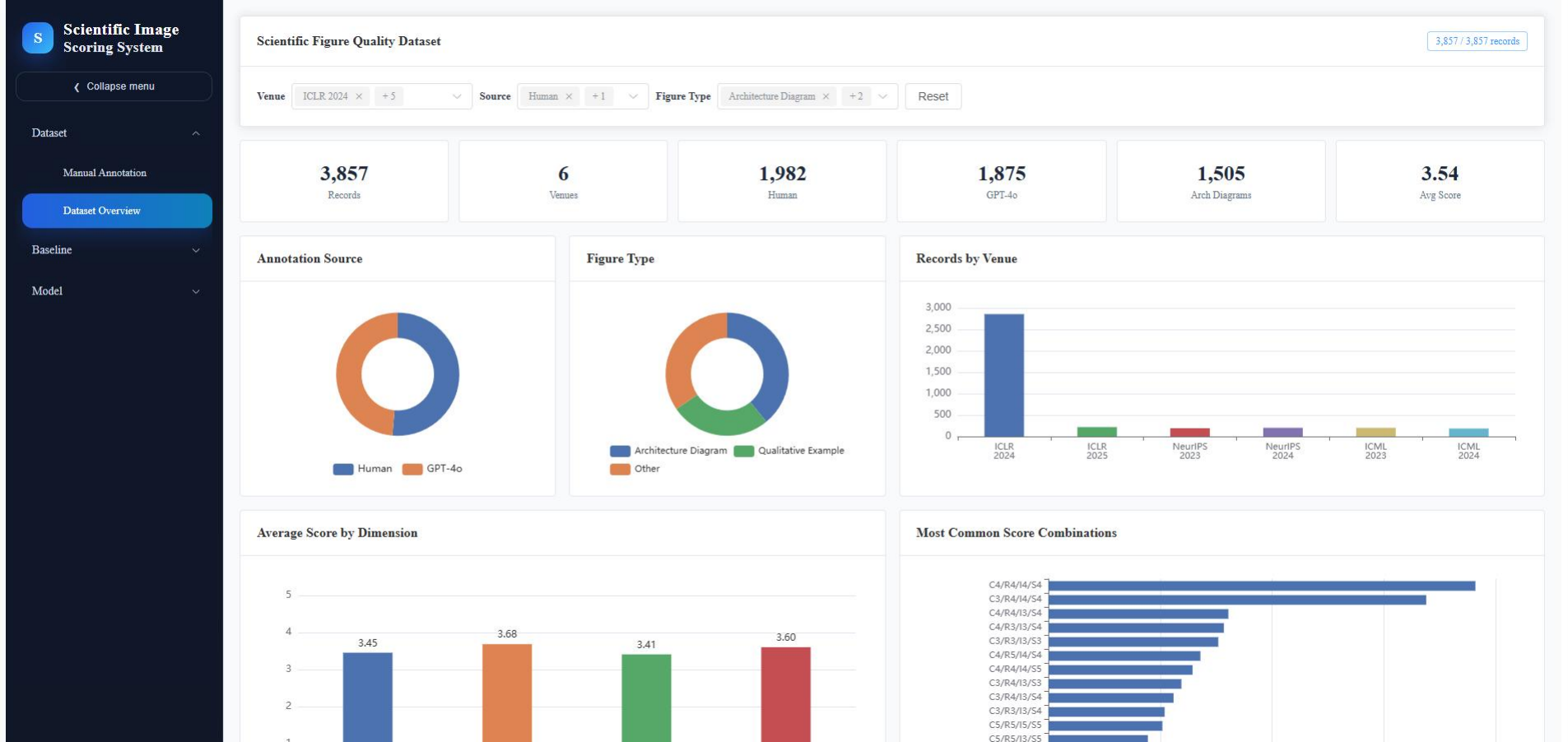}
  \caption{Dataset Dashboard: 3,857-figure overview with venue/source/type filters, annotation and type mixes, venue counts, and per-dimension averages (mean overall $3.54$).}
  \label{fig:app-dataset-dash}
\end{figure*}

\section{Explainability and Dashboards}
\label{app:explain}

After fusion, CubeMLP gates produce modality weights over $\{I,c,\mathcal{X},a,m\}$.
On representative test instances, context and abstract are each near $0.21$, while the image stream is near $0.18$.
Cross-attention maps typically highlight axis labels, method boxes, and panel titles rather than whitespace; attention entropy is relatively stable in late training epochs.
These signals are meant for inspection only and do not replace human review.

\begin{figure*}[t]
  \centering
  \includegraphics[width=\textwidth]{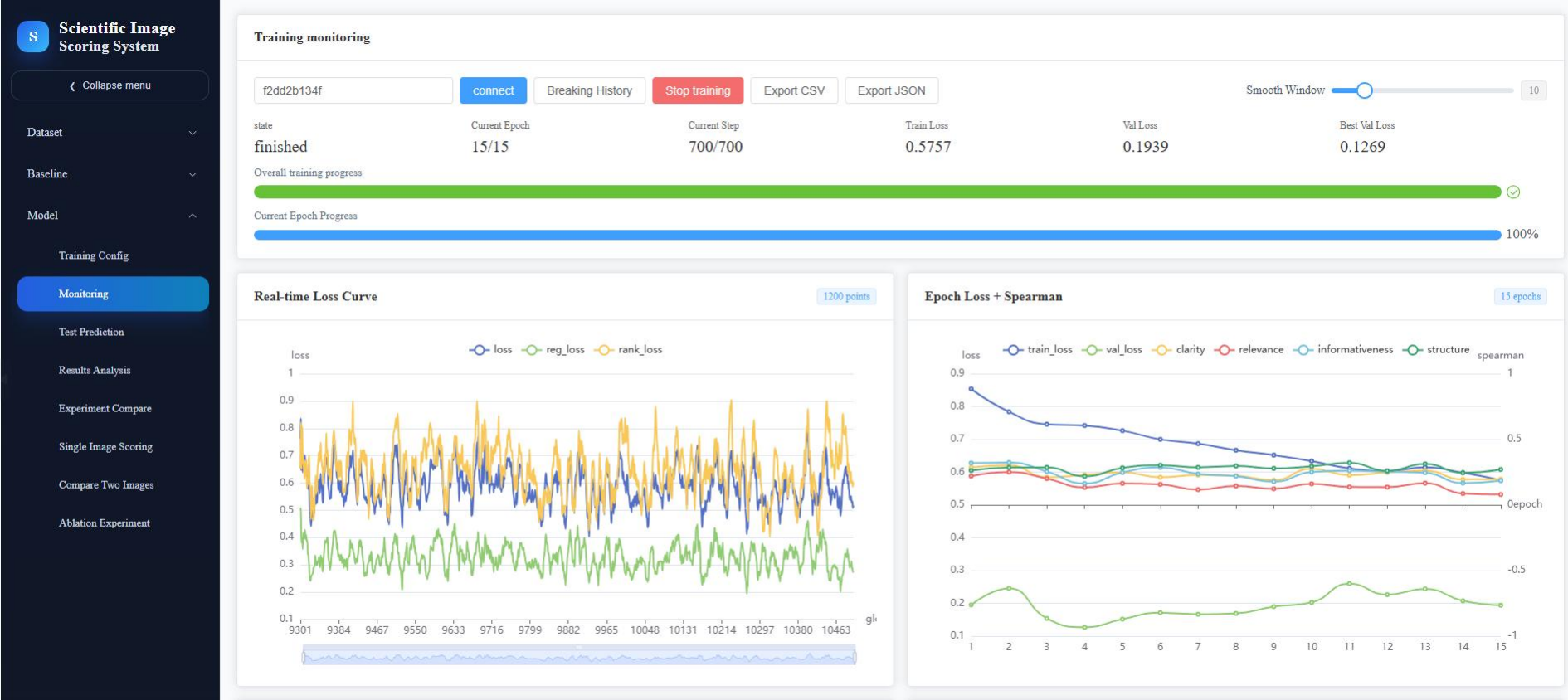}
  \caption{Model Monitoring Dashboard: training status with regression/ranking losses and per-dimension Spearman curves.}
  \label{fig:app-monitor-dash}
\end{figure*}

\begin{figure*}[t]
  \centering
  \includegraphics[width=\textwidth]{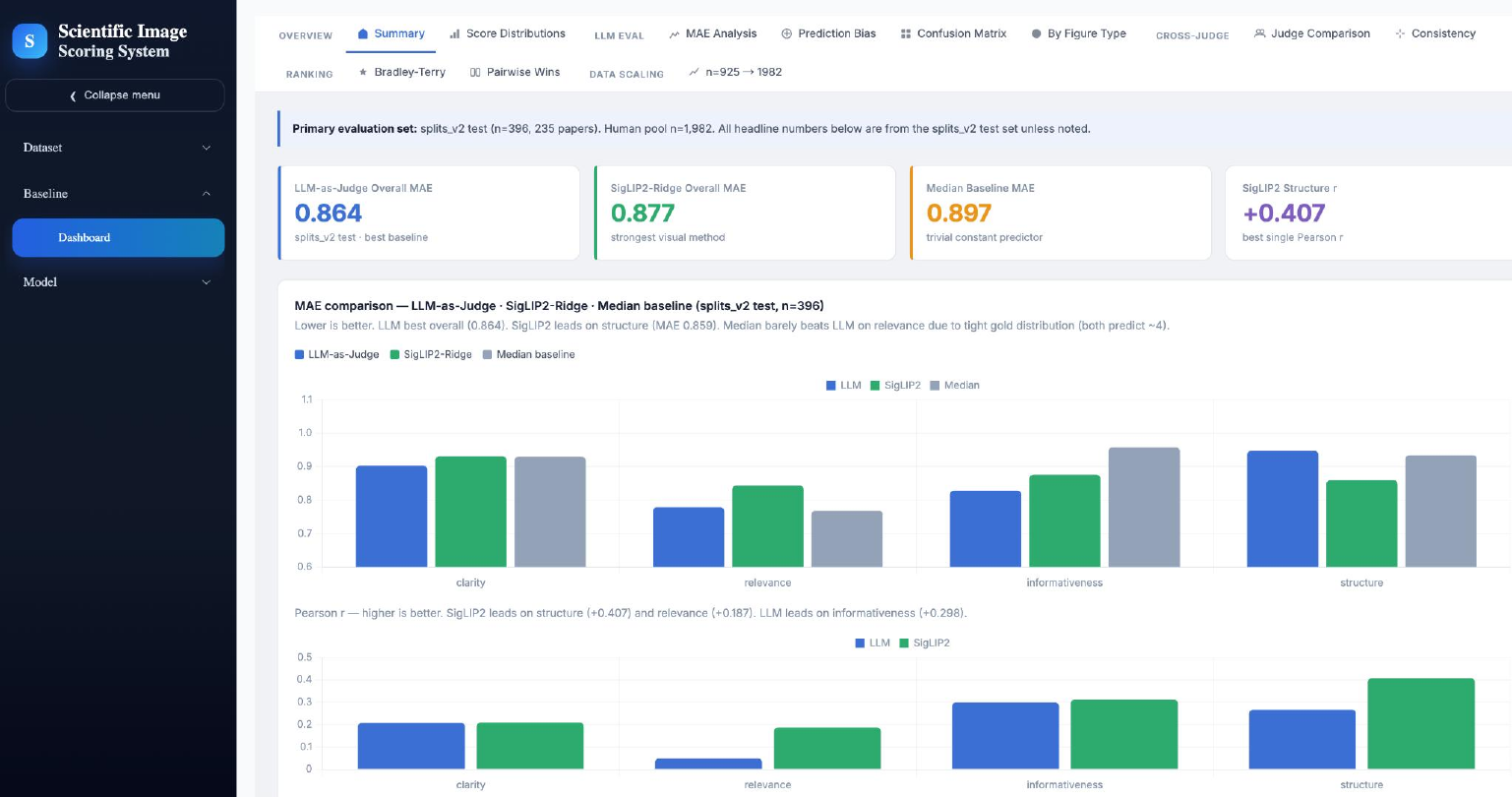}
  \caption{Baseline Dashboard: interactive comparison of constant, similarity, ridge, and LLM-judge baselines.
    Top cards report \emph{evaluation} metrics on the held-out split---macro MAE (absolute error on the 1--5 rubric, lower is better) and per-dimension Pearson~$r$ (correlation between predictions and human gold; the leading ``$+$'' marks a positive $r$, e.g.\ $+0.407$ on Structure).
    These $0.xx$ values are \emph{not} normalized per-dimension scores.}
  \label{fig:app-baseline-dash}
\end{figure*}

\begin{figure*}[t]
  \centering
  \includegraphics[width=\textwidth]{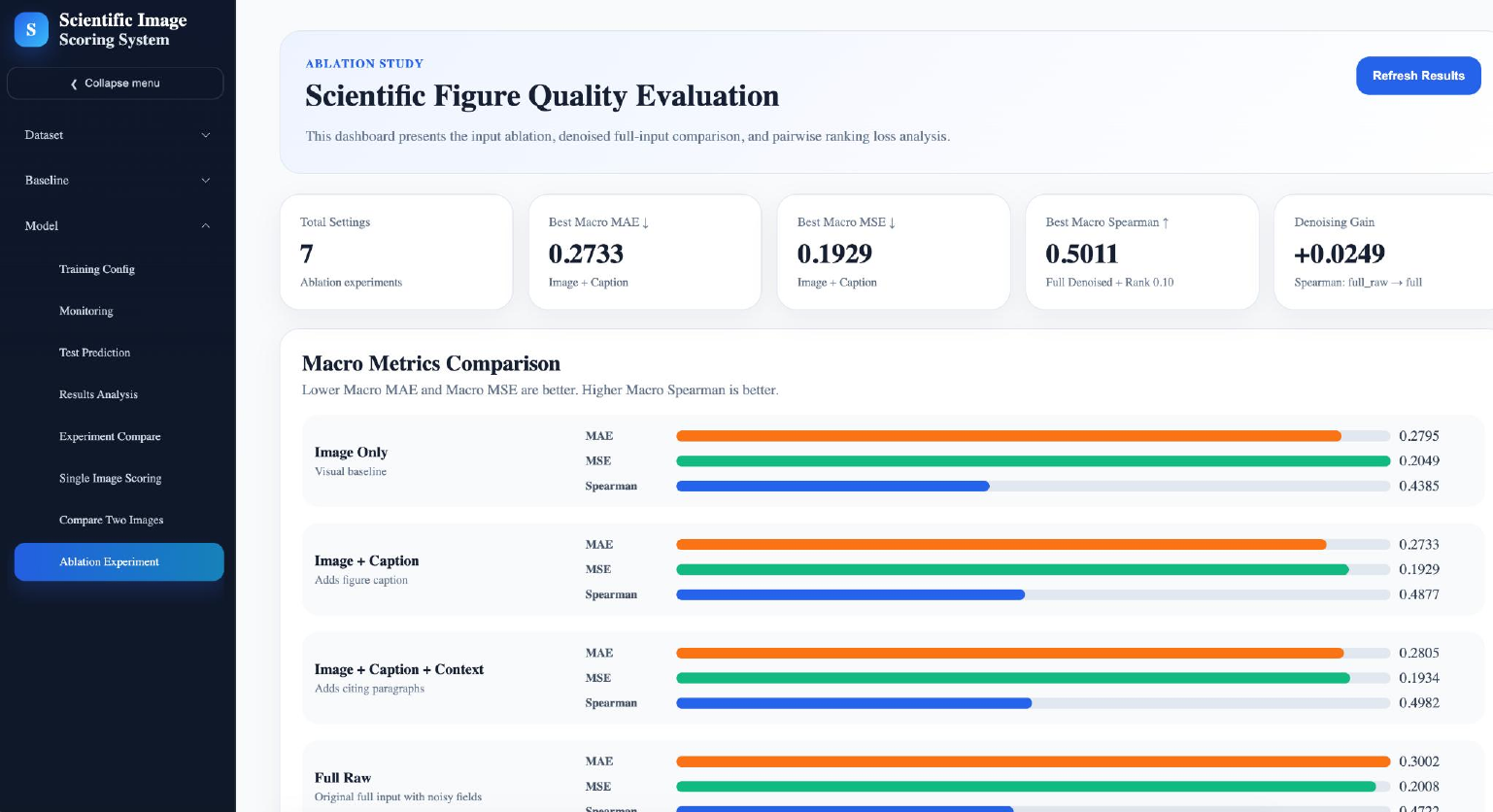}
  \caption{Ablation Studies Dashboard: interactive view of input/denoising/ranking configurations.}
  \label{fig:app-ablation-dash}
\end{figure*}

Separately, we built a local development dashboard for corpus browsing and training monitoring (not a public cloud service).
It supports baseline/ablation comparison, single-figure and pairwise scoring, and CSV/JSON export.
Figures~\ref{fig:app-dataset-dash}--\ref{fig:app-ablation-dash} are screenshots of that tool; reproducing the main tables does not require the UI.

\subsection{Human vs.\ Model Scoring Cases}
\label{app:scoring-cases}

\begin{figure*}[p]
  \centering
  \includegraphics[width=\textwidth,height=0.88\textheight,keepaspectratio]{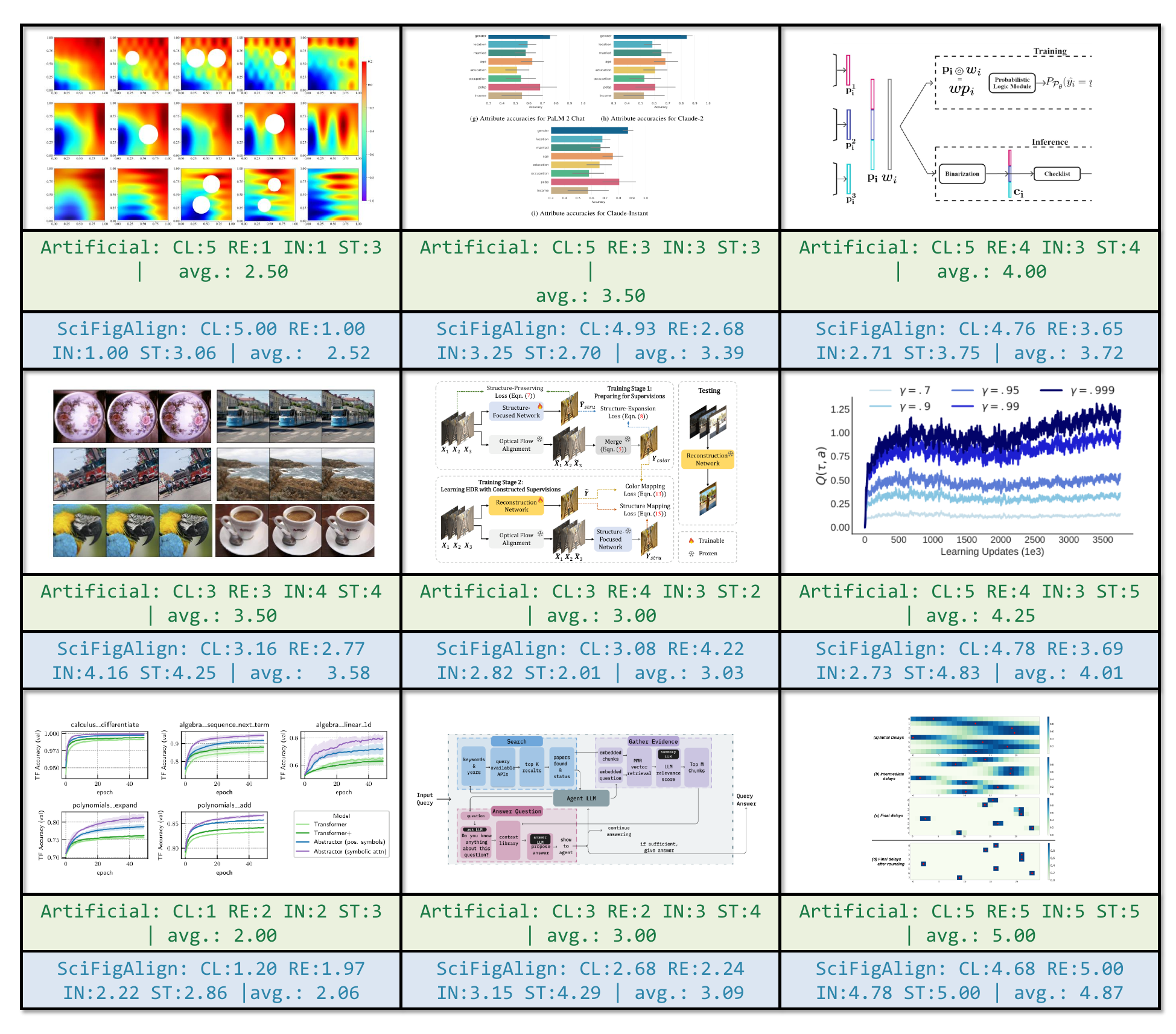}
  \caption{Human--SciFigAlign scoring comparison, \textbf{case batch~1} (9 representative figures; 3-column grid).
    All four dimensions (CL/RE/IN/ST) are on the 1--5 rubric scale (two-decimal display; same units as training labels, not MAE or Pearson~$r$).}
  \label{fig:app-scoring-case1}
\end{figure*}

\begin{figure*}[p]
  \centering
  \includegraphics[width=\textwidth,height=0.88\textheight,keepaspectratio]{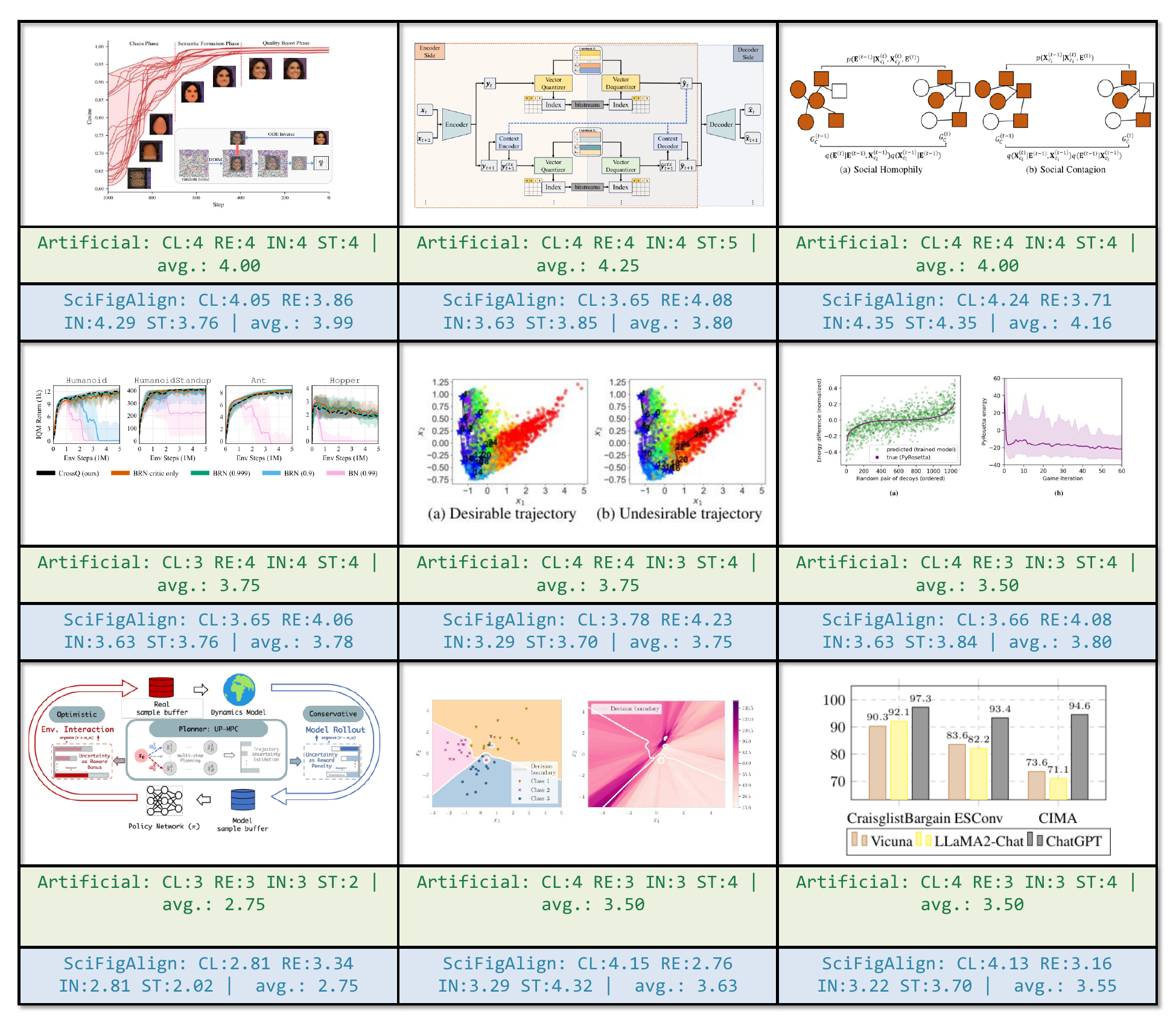}
  \caption{Human--SciFigAlign scoring comparison, \textbf{case batch~2} (9 representative figures; 3-column grid).
    Same display convention as Figure~\ref{fig:app-scoring-case1}.}
  \label{fig:app-scoring-case2}
\end{figure*}

Figures~\ref{fig:app-scoring-case1} and~\ref{fig:app-scoring-case2} compare human gold scores and SciFigAlign predictions on 18 evaluation figures (two batches of nine).
Each cell shows the figure crop with Clarity, Relevance, Informativeness, and Structure on the same 1--5 rubric used in training (human scores are integers; model scores are continuous).
As in \S\ref{app:arch-heads}, each head uses a sigmoid scaled by~5, so predictions lie in $(0,5)$.
Green rows are human gold; blue rows are model predictions.
These grids illustrate typical agreement patterns and do not replace the held-out test-set metrics.
Note that Figure~\ref{fig:app-baseline-dash} plots evaluation diagnostics (MAE and Pearson~$r$), not raw CL/RE/IN/ST scores.

\begin{table}[t]
  \centering
  \small
  \caption{Validation ablation suite (seven settings; not equated to the test ladder).}
  \label{tab:val-ablation}
  \setlength{\tabcolsep}{6pt}
  \begin{tabular}{@{}lccc@{}}
    \toprule
    Setting & MAE$\downarrow$ & MSE$\downarrow$ & SRCC$\uparrow$ \\
    \midrule
    Image only & 0.2795 & 0.2049 & 0.4385 \\
    Image + caption & 0.2733 & 0.1929 & 0.4877 \\
    + context & 0.2805 & 0.1934 & 0.4982 \\
    Full raw & 0.3002 & 0.2008 & 0.4722 \\
    Full denoised & 0.2797 & 0.1956 & 0.4970 \\
    + rank $\lambda{=}0.05$ & 0.2794 & 0.1952 & 0.4994 \\
    + rank $\lambda{=}0.10$ & 0.2796 & 0.1952 & 0.5011 \\
    \bottomrule
  \end{tabular}
\end{table}

\section{Training Dynamics, Corpus, and Paradigm}
\label{app:paradigm}

\noindent\textbf{Training dynamics.}
Figure~\ref{fig:app-monitor-dash} shows live training curves.
Train and validation losses decrease without ranking-loss divergence, and Spearman correlation rises over epochs (Clarity/Structure typically improve earlier than Relevance/Informativeness).
Cross-attention entropy stays stable in late epochs.

\noindent\textbf{Modality contribution.}
Qualitatively, the image stream contributes most to Clarity; citing context and metadata matter more for Relevance; captions contribute strongly to Informativeness; Structure uses layout cues together with abstract-level narrative (the image stream remains necessary).
This is consistent with Appendix~F: Image+Caption is already strong on MAE, while full denoised inputs plus ranking favour Spearman / within-paper ordering.

\noindent\textbf{Corpus mix.}
The cleaned corpus has 3,857 figures from 3,126 papers (from 12,492 raw crops), with type mix architecture 39.0\% / other 34.6\% / qualitative 26.4\%, annotation mix human 1,982 / GPT-4o-assisted 1,875, mean overall score $3.54$, and context-field coverage ${>}99\%$.

\noindent\textbf{Paradigm note.}
SciFigAlign differs from SIQA-style prompting and black-box LLM judges by using a fine-tuned CrossAttn+CubeMLP regressor with exportable attention/weights and an explicit pairwise hinge.
Main claims are restricted to this ICLR/NeurIPS/ICML corpus and the held-out human-rated evaluation protocol.

\section{Extended Ablations, Judge Modes, and Failures}
\label{app:extra2}

Table~\ref{tab:val-ablation} expands the matched \emph{validation} protocol from the main paper.
These validation numbers are \emph{not} the same as the \emph{test} ladder
$0.890\to0.720\to0.650\to0.580\to0.520\to0.3524$.
On validation, Image+Caption achieves the lowest Macro MAE/MSE ($0.2733$ / $0.1929$), while Full Denoised$+\lambda{=}0.10$ achieves the highest Macro Spearman ($0.5011$).
Full Raw is weakest on absolute error (MAE $0.3002$); denoising improves MAE/MSE/SRCC over Full Raw.

\noindent\textbf{Test ladder and judge modes.}
On the held-out human-rated test subset: Image only $0.890$ $\to$ +caption $0.720$ $\to$ +context $0.650$ $\to$ Full raw $0.580$ $\to$ Full denoised $0.520$ $\to$ Full+rank $0.3524$ (PA $81.64\%$; SRCC $0.3088$; $|\mathcal{P}|{=}365$).
On a 200-figure subset, GPT-5.4-mini full-context MAE $0.778$ beats Gemini-2.5-Flash $0.874$, but remains close to an always-4 constant predictor ($0.800$).
Input-mode judge variants still underperform SciFigAlign on the held-out test set.

\noindent\textbf{Failure modes.}
OCR-dense plots, crowded multi-panel grids, and Informativeness disagreements (sparse visual + strong caption claim) remain the hardest cases.
Stronger OCR/layout features for the visual stream are a natural next step.

\noindent\textbf{Reproducibility.}
Supplementary material provides paper-level splits, instance fields $(I,c,\mathcal{X},a,m)$, Table~\ref{tab:hparams}, and MAE/SRCC/PA evaluation scripts (see project page).
Hardware: a single consumer GPU with 8\,GB memory, PyTorch~$\ge$2.0, FP16 (${\approx}45$\,min for 5 epochs on 3k+ samples).
Baselines share the same instance fields but not SciFigAlign weights.
PA uses $|\mathcal{P}|{=}365$ pairs with gap $\ge0.5$; bootstrap MAE CI $[0.331,0.374]$ vs.\ LLM $[0.841,0.887]$.

\section{Limitations and Scope of Claims}
\label{app:limitations}

\noindent\textbf{Annotation and corpus scope.}
We use mixed human and GPT-4o labels (51.4\% / 48.6\%) rather than an all-expert corpus.
The benchmark covers ICLR, NeurIPS, and ICML figures only; biology, chemistry, medicine, and non-English venues are not represented.
We do not report human--human agreement ceilings on this corpus; multi-rater agreement on a held-out subset and broader venue coverage remain open.

\noindent\textbf{Baselines and failure modes.}
LLM-judge baselines depend on proprietary APIs and prompt engineering, which limits exact reproducibility and cost-controlled replication.
SciFigAlign still errs on OCR-dense charts, crowded multi-panel grids, and subjective Informativeness cases where sparse visuals meet strong caption claims.
Domain shift to wet-lab experimental figures, denser panel layouts, and non-English manuscripts is untested.

\noindent\textbf{What this supplement covers.}
Rubric design, corpus construction, hyperparameters, explainability tooling, extended ablations, judge protocols, and related-work task comparisons appear in Appendices~A--F.
Claims in the main paper should be read within these scopes rather than as universal scientific-figure assessment standards.

\end{document}